\documentclass{clv3}

\usepackage{hyperref}
\usepackage{xcolor}
\usepackage{microtype}
\usepackage{url}
\usepackage{latexsym}
\usepackage{amsthm}
\usepackage{amssymb}
\usepackage{amsfonts}
\usepackage{amsmath}
\usepackage{microtype}
\usepackage{adjustbox}
\usepackage{tabularx}
\usepackage{booktabs}
\usepackage{multirow}
\usepackage{mdwlist}
\usepackage{colortbl}
\usepackage{subcaption}
\usepackage{makecell}
\usepackage{arydshln}
\usepackage{bm}
\usepackage{CJKutf8} 
\newcolumntype{Y}{>{\centering\arraybackslash}X}
\newcolumntype{P}{>{\raggedleft\arraybackslash}p{.235in}}

\usepackage[colorinlistoftodos, textsize=footnotesize]{todonotes}

\definecolor{darkblue}{rgb}{0, 0, 0.5}
\definecolor{Gray}{gray}{0.92}
\definecolor{applegreen}{rgb}{0.55, 0.71, 0.0}

\hypersetup{colorlinks=true,citecolor=darkblue, linkcolor=darkblue, urlcolor=darkblue}
\issue{1}{1}{2020}

\runningtitle{Multi-SimLex}

\runningauthor{Vuli\'{c} et al.}

\begin{document}

\title{Multi-SimLex: A Large-Scale Evaluation of Multilingual and Cross-Lingual Lexical Semantic Similarity \\
{\large\url{https://multisimlex.com/}}
}

\author{Ivan Vuli\'{c} \thanks{$^{\spadesuit}$\textit{Equal contribution}; English Faculty Building, 9 West Road Cambridge CB3 9DA, United Kingdom. E-mail: \texttt{\{iv250,sb895,ep490,om304,alk23\}@cam.ac.uk}}$^{\spadesuit}$}
\affil{LTL, University of Cambridge}

\author{Simon Baker $^{*\spadesuit}$}
\affil{LTL, University of Cambridge}

\author{Edoardo Maria Ponti $^{*\spadesuit}$}
\affil{LTL, University of Cambridge}

\author{Ulla Petti $^{*}$}
\affil{LTL, University of Cambridge}

\author{Ira Leviant \thanks{Technion City, Haifa 3200003, Israel. E-mail: \texttt{\{ira.leviant,edenb\}@campus.technion.ac.il, roiri@ie.technion.ac.il}}}
\affil{Faculty of Industrial Engineering and Management, Technion, IIT}

\author{Kelly Wing $^{*}$}
\affil{LTL, University of Cambridge}

\author{Olga Majewska $^{*}$}
\affil{LTL, University of Cambridge}

\author{Eden Bar $^{**}$}
\affil{Faculty of Industrial Engineering and Management, Technion, IIT}

\author{Matt Malone $^{*}$}
\affil{LTL, University of Cambridge}

\author{Thierry Poibeau \thanks{ Rue Maurice Arnoux, 92120 Montrouge, France. E-mail: \texttt{thierry.poibeau@ens.fr}}}
\affil{LATTICE Lab, CNRS and ENS/PSL and Univ. Sorbonne nouvelle/USPC}

\author{Roi Reichart $^{**}$}
\affil{Faculty of Industrial Engineering and Management, Technion, IIT}

\author{Anna Korhonen $^{*}$}
\affil{LTL, University of Cambridge}

\maketitle

\begin{abstract}
We introduce Multi-SimLex, a large-scale lexical resource and evaluation benchmark covering datasets for 12 typologically diverse languages, including major languages (e.g., Mandarin Chinese, Spanish, Russian) as well as less-resourced ones (e.g., Welsh, Kiswahili). Each language dataset is annotated for the lexical relation of semantic similarity and contains 1,888 semantically aligned concept pairs, providing a representative coverage of word classes (nouns, verbs, adjectives, adverbs), frequency ranks, similarity intervals, lexical fields, and concreteness levels. 
Additionally, owing to the alignment of concepts across languages, we provide a suite of 66 cross-lingual semantic similarity datasets.
Due to its extensive size and language coverage, Multi-SimLex provides entirely novel opportunities for experimental evaluation and analysis.
On its monolingual and cross-lingual benchmarks, we evaluate and analyze a wide array of recent state-of-the-art monolingual and cross-lingual representation models, including static and contextualized word embeddings (such as fastText, M-BERT and XLM), 
externally informed lexical representations, as well as fully unsupervised and (weakly) supervised cross-lingual word embeddings. 
We also present a step-by-step dataset creation protocol for creating consistent, Multi-Simlex -style resources  for additional languages. 
We make these contributions - the public release of Multi-SimLex datasets, their creation protocol, strong baseline results, 
and in-depth analyses which can be be helpful in guiding future developments in multilingual lexical semantics and representation learning - available via a website which will encourage community effort in further expansion of Multi-Simlex to many more languages. Such a large-scale semantic resource could inspire significant further advances in NLP across languages.
\end{abstract}

\section{Introduction}
\label{s:introduction}
The lack of annotated training and evaluation data for many tasks and domains hinders the development of computational models for the majority of the world's languages \cite{Snyder:2010icml,Adams:2017eacl,Ponti:2019cl}. The necessity to guide and advance multilingual and cross-lingual NLP through annotation efforts that follow cross-lingually consistent guidelines has been recently recognized by collaborative initiatives such as the Universal Dependency (UD) project \cite{UD2.4}. The latest version of UD (as of March 2020) covers more than 70 languages. Crucially, this resource continues to steadily grow and evolve through the contributions of annotators from across the world, extending the UD's reach to a wide array of typologically diverse languages. Besides steering research in multilingual parsing \cite{Zeman:2018conll,Kondratyuk:2019emnlp,Doitch:2019tacl} and cross-lingual parser transfer \cite{Rasooli:2017tacl,Lin:2019acl,Rotman:2019tacl}, the consistent annotations and guidelines have also enabled a range of insightful comparative studies focused on the languages' syntactic (dis)similarities \cite{Bjerva:2018naacl,Ponti:2018acl,Pires:2019acl}. 

Inspired by the UD work and its substantial impact on research in (multilingual) syntax, in this article we introduce \textbf{Multi-SimLex}, a suite of manually and consistently annotated \textbf{semantic datasets} for 12 different languages, focused on the fundamental lexical relation of \textbf{semantic similarity} \cite{budanitsky2006evaluating,Hill:2015cl}. 
For any pair of words, this relation measures whether their referents share the same (functional) features, as opposed to general cognitive association captured by co-occurrence patterns in texts (i.e., the distributional information). 
Datasets that quantify the strength of true semantic similarity between concept pairs such as SimLex-999 \cite{Hill:2015cl} or SimVerb-3500 \cite{Gerz:2016emnlp} have been instrumental in improving models for distributional semantics and representation learning. Discerning between semantic similarity and relatedness/association is not only crucial for theoretical studies on lexical semantics (see \S\ref{s:similarity}), but has also been shown to benefit a range of language understanding tasks in NLP. Examples include dialog state tracking \cite{Mrksic:2017tacl,Ren:2018emnlp}, spoken language understanding \cite{Kim:2016slt,Kim:16b}, text simplification \cite{Glavas:2018acl,Ponti:2018emnlp,lauscher2019informing}, dictionary and thesaurus construction \cite{Cimiano:2005jair,Hill:2016tacl}.

Despite the proven usefulness of semantic similarity datasets,
they are available only for a small and typologically narrow sample of resource-rich languages such as German, Italian, and Russian \cite{Leviant:2015arxiv}, whereas some language types and low-resource languages typically lack similar evaluation data. Even if some resources do exist, they are limited in their \textit{size} (e.g., 500 pairs in Turkish \cite{Ercan:2018coling}, 500 in Farsi \cite{Camacho:2017semeval}, or 300 in Finnish \cite{Venekoski:2017nodalida}) and \textit{coverage} (e.g., all datasets which originated from the original English SimLex-999 contain only high-frequent concepts, and are dominated by nouns). This is why, as our departure point, we introduce a \textbf{larger and more comprehensive} English word similarity dataset spanning 1,888 concept pairs (see \S\ref{s:ensimlex}).


Most importantly, semantic similarity datasets in different languages have been created using heterogeneous construction procedures with different guidelines for translation and annotation, as well as different rating scales. For instance, some datasets were obtained by directly translating the English SimLex-999 in its entirety \cite{Leviant:2015arxiv,Mrksic:2017tacl} or in part \cite{Venekoski:2017nodalida}. Other datasets were created from scratch \cite{Ercan:2018coling} and yet others sampled English concept pairs differently from SimLex-999 and then translated and reannotated them in target languages \cite{Camacho:2017semeval}. This heterogeneity makes these datasets incomparable and precludes systematic cross-linguistic analyses. In this article, consolidating the lessons learned from previous dataset construction paradigms, we propose a carefully designed \textbf{translation and annotation protocol} for developing monolingual Multi-SimLex datasets with aligned concept pairs for 
typologically diverse languages. We apply this protocol to a set of 12 languages, including a mixture of major languages (e.g., Mandarin, Russian, and French) as well as several low-resource ones (e.g., Kiswahili, Welsh, and Yue Chinese). We demonstrate that our proposed dataset creation procedure yields data with high inter-annotator agreement rates (e.g., the average mean inter-annotator agreement for Welsh is 0.742).


The unified construction protocol and alignment between concept pairs enables a series of quantitative analyses. Preliminary studies on the influence that polysemy and cross-lingual variation in lexical categories (see \S\ref{ss:semtyp}) have on similarity judgments 
are provided in \S\ref{s:multisimlex}. Data created according to Multi-SimLex protocol also allow for probing into whether similarity judgments are universal across languages, or rather depend on linguistic affinity (in terms of linguistic features, phylogeny, and geographical location). We investigate this question in \S\ref{ss:typsim}. Naturally, Multi-SimLex datasets can be used as an intrinsic evaluation benchmark to assess the quality of lexical representations based on monolingual, joint multilingual, and transfer learning paradigms. We conduct a systematic evaluation of several state-of-the-art representation models in \S\ref{s:monoeval}, showing that there are large gaps between human and system performance in all languages. The proposed construction paradigm also supports the automatic creation of 66 cross-lingual Multi-SimLex datasets by interleaving the monolingual ones. 
We outline the construction of the cross-lingual datasets in \S\ref{s:xsimlex}, and then present a quantitative evaluation of a series of cutting-edge cross-lingual representation models on this benchmark in \S\ref{ss:xling-eval}.

\vspace{2mm}
\noindent \textit{Contributions.}
We now summarize the main contributions of this work:
\vspace{1.6mm}

\noindent 1) Building on lessons learned from prior work, we create a more comprehensive lexical semantic similarity dataset for the English language spanning a total of 1,888 concept pairs balanced with respect to similarity, frequency, and concreteness, and covering four word classes: nouns, verbs, adjectives and, for the first time, adverbs. This dataset serves as the main source for the creation of equivalent datasets in several other languages.

\vspace{1.4mm}
\noindent 2) We present a carefully designed and rigorous language-agnostic translation and annotation protocol. These well-defined guidelines will facilitate the development of future Multi-SimLex datasets for other languages. The proposed protocol eliminates some crucial issues with prior efforts focused on the creation of multi-lingual semantic resources, namely: i) limited coverage; ii) heterogeneous annotation guidelines; and iii) concept pairs which are semantically incomparable across different languages. 

\vspace{1.4mm}
\noindent 3) We offer to the community manually annotated evaluation sets of 1,888 concept pairs across 12 typologically diverse languages, and 66 large cross-lingual evaluation sets. To the best of our knowledge, Multi-SimLex is the most comprehensive evaluation resource to date focused on the relation of semantic similarity. 

\vspace{1.4mm}
\noindent 4) We benchmark a wide array of recent state-of-the-art monolingual and cross-lingual word representation models across our sample of languages. The results can serve as strong baselines that lay the foundation for future improvements.

\vspace{1.4mm}
\noindent 5) We present a first large-scale evaluation study on the ability of encoders pretrained on language modeling (such as \textsc{bert} \cite{devlin2018bert} and \textsc{xlm} \cite{Conneau:2019nips}) to reason over word-level semantic similarity in different languages. To our own surprise, the results show that  monolingual pretrained encoders, even when presented with word types out of context, are sometimes competitive with static word embedding models such as fastText \cite{Bojanowski:2017tacl} or \texttt{word2vec} \cite{Mikolov:2013nips}. The results also reveal a huge gap in performance between massively multilingual pretrained encoders and language-specific encoders in favor of the latter: our findings support other recent empirical evidence related to the ``curse of multilinguality'' \cite{Conneau:2019arxiv,Bapna:2019emnlp} in representation learning. 

\vspace{1.4mm}
\noindent 6) We make all of these resources available on a website which facilitates easy creation, submission and sharing of Multi-Simlex-style datasets for a larger number of languages. We hope that this will yield an even larger repository of semantic resources that inspire future advances in NLP within and across languages.


\vspace{1.6mm}
In light of the success of Universal Dependencies \cite{UD2.4}, we hope that our initiative will instigate a collaborative public effort with established and clear-cut guidelines that will result in additional Multi-SimLex datasets in a large number of languages in the near future. 
Moreover, we hope that it will provide means to advance our understanding of distributional and lexical semantics across a large number of languages. 
All monolingual and cross-lingual Multi-SimLex datasets--along with detailed translation and annotation guidelines--are available online at: \url{https://multisimlex.com/}.

\section{Lexical Semantic Similarity}
\label{s:similarity}

\subsection{Similarity and Association}
\label{ss:sim-vs-assoc}
The focus of the Multi-SimLex initiative is on the lexical relation of pure \textit{semantic similarity}. For any pair of words, this relation measures whether their referents share the same features. For instance, \textit{graffiti} and \textit{frescos} are similar to the extent that they are both forms of painting and appear on walls. This relation can be contrasted with the cognitive \textit{association} between two words, which often depends on how much their referents interact in the real world, or are found in the same situations. For instance, a \textit{painter} is easily associated with \textit{frescos}, although they lack any physical commonalities. Association is also known in the literature under other names: relatedness \cite{budanitsky2006evaluating}, topical similarity \citep{hatzivassiloglou2001simfinder}, and domain similarity \citep{turney2012domain}.

Semantic similarity and association overlap to some degree, but do not coincide \cite{Kiela:2015emnlp,Vulic:2017eacleval}. In fact, there exist plenty of pairs that are intuitively associated but not similar. Pairs where the converse is true can also be encountered, although more rarely. An example are synonyms where a word is common and the other infrequent, such as \textit{to seize} and \textit{to commandeer}. \citet{Hill:2015cl} revealed that while similarity measures based on the WordNet graph \citep{wu1994verb} and human judgments of association in the University of South Florida Free Association Database \citep{Nelson:2004usf} do correlate, a number of pairs follow opposite trends. 
Several studies on human cognition also point in the same direction. For instance, semantic priming can be triggered by similar words without association \citep{lucas2000semantic}. On the other hand, a connection with cue words is established more quickly for topically related words rather than for similar words in free association tasks \cite{de2008word}.

A key property of semantic similarity is its \textit{gradience}: pairs of words can be similar to a different degree. On the other hand, the relation of  \textit{synonymy} is binary: pairs of words are synonyms if they can be substituted in all contexts (or most contexts, in a looser sense), otherwise they are not. While synonyms can be conceived as lying on one extreme of the semantic similarity continuum, it is crucial to note that their definition is stated in purely relational terms, rather than invoking their referential properties \citep{lyons1977semantics,cruse1986lexical,coseriu1967lexikalische}. This makes behavioral studies on semantic similarity fundamentally different from lexical resources like WordNet \cite{Miller:95}, which include paradigmatic relations (such as synonymy).

\subsection{Similarity for NLP: Intrinsic Evaluation and Semantic Specialization}

The ramifications of the distinction between similarity and association are profound for distributional semantics. This paradigm of lexical semantics is grounded in the distributional hypothesis, formulated by \citet{firth1957synopsis} and \citet{harris1951methods}. According to this hypothesis, the meaning of a word can be recovered empirically from the contexts in which it occurs within a collection of texts. Since both pairs of topically related words and pairs of purely similar words tend to appear in the same contexts, their associated meaning confounds the two distinct relations \cite{Hill:2015cl,Schwartz:2015conll,Vulic:2017conll}. As a result, distributional methods obscure a crucial facet of lexical meaning. 

This limitation also reflects onto word embeddings (WEs), representations of words as low-dimensional vectors that have become indispensable for a wide range of NLP applications \cite[\textit{inter alia}]{Collobert:2011jmlr,Chen:2014emnlp,Melamud:2016naacl}. In particular, it involves both \textit{static} WEs learned from co-occurrence patterns \cite{Mikolov:2013nips,Levy:2014acl,Bojanowski:2017tacl} and \textit{contextualized} WEs learned from modeling word sequences \citep[\textit{inter alia}]{Peters:2018naacl,devlin2018bert}. As a result, in the induced representations, geometrical closeness (measured e.g.\ through cosine distance) conflates genuine similarity with broad relatedness. For instance, the vectors for antonyms such as \textit{sober} and \textit{drunk}, by definition dissimilar, might be neighbors in the semantic space under the distributional hypothesis. \citet{turney2012domain}, \citet{kiela2014systematic}, and \citet{Melamud:2016naacl} demonstrated that different choices of hyper-parameters in WE algorithms (such as context window) emphasize different relations in the resulting representations. Likewise, \citet{agirre2009study} and \citet{Levy:2014acl} discovered that WEs learned from texts annotated with syntactic information mirror similarity better than simple local bag-of-words neighborhoods.

The failure of WEs to capture semantic similarity, in turn, affects model performance in several NLP applications where such knowledge is crucial. In particular, Natural Language Understanding tasks such as statistical dialog modeling, text simplification, or semantic text similarity \cite{Mrksic:2016naacl,Kim:2016slt,Ponti:2019emnlp}, among others, suffer the most. As a consequence, resources providing clean information on semantic similarity are key in mitigating the side effects of the distributional signal. In particular, such databases can be employed for the \textit{intrinsic evaluations} of specific WE models as a proxy of their reliability for downstream applications \citep{collobert2008unified,baroni2010distributional,Hill:2015cl}; intuitively, the more WEs are misaligned with human judgments of similarity, the more their performance on actual tasks is expected to be degraded. Moreover, word representations can be \textit{specialized} (a.k.a.\ retrofitted) by disentangling word relations of similarity and association. In particular, linguistic constraints sourced from external databases (such as synonyms from WordNet) can be injected into WEs \cite[\textit{inter alia}]{Faruqui:2015naacl,Wieting:2015tacl,Mrksic:2017tacl,lauscher2019informing,Kamath:2019ws} in order to enforce a particular relation in a distributional semantic space while preserving the original adjacency properties.

\subsection{Similarity and Language Variation: Semantic Typology}
\label{ss:semtyp}
In this work, we tackle the concept of (true) semantic similarity from a multilingual perspective. While the same meaning representations may be shared by all human speakers at a deep cognitive level, there is no one-to-one mapping between the words in the lexicons of different languages. This makes the comparison of similarity judgments across languages difficult, since the meaning overlap of translationally equivalent words is sometimes far less than exact. This results from the fact that the way languages `partition' semantic fields is partially arbitrary \citep{trier1931deutsche}, although constrained cross-lingually by common cognitive biases \cite{majid2007semantic}. For instance, consider the field of colors: English distinguishes between \textit{green} and \textit{blue}, whereas Murle (South Sudan) has a single word for both \citep{wals-134}.

In general, \textit{semantic typology} studies the variation in lexical semantics across the world's languages. According to \citep{Evans-2011}, the ways languages categorize concepts into the lexicon follow three main axes: 1) \textit{granularity}: what is the number of categories in a specific domain?; 2) \textit{boundary location}: where do the lines marking different categories lie?; 3) \textit{grouping and dissection}: what are the membership criteria 
of a category; which instances are considered to be more prototypical? Different choices with respect to these axes lead to different lexicalization patterns.\footnote{More formally, \textit{colexification} is a phenomenon when different meanings can be expressed by the same word in a language \cite{Francois:2008colex}. For instance, the two senses which are distinguished in English as \textit{time} and \textit{weather} are co-lexified in Croatian: the word \textit{vrijeme} is used in both cases.} For instance, distinct senses in a polysemous word in English, such as \textit{skin} (referring to both the body and fruit), may be assigned separate words in other languages such as Italian \textit{pelle} and \textit{buccia}, respectively \citep{rzymski2020database}. We later analyze whether similarity scores obtained from native speakers also loosely follow the patterns described by semantic typology.

\section{Previous Work and Evaluation Data}
\label{s:rw}
\noindent \textit{Word Pair Datasets.} Rich expert-created resources such as WordNet \cite{Miller:95,Fellbaum:1998wn}, VerbNet \cite{Kipper:2005thesis,Kipper:2008lre}, or FrameNet \cite{Baker:98} encode a wealth of semantic and syntactic information, but are expensive and time-consuming to create. The scale of this problem gets multiplied by the number of languages in consideration. Therefore, crowd-sourcing with non-expert annotators has been adopted as a quicker alternative to produce smaller and more focused semantic resources and evaluation benchmarks. This alternative practice has had a profound impact on distributional semantics and representation learning \cite{Hill:2015cl}. While some prominent English word pair datasets such as WordSim-353 \cite{Finkelstein:2002tois}, MEN \cite{Bruni:2014jair}, or Stanford Rare Words \cite{Luong:2013conll} did not discriminate between similarity and relatedness, the importance of this distinction was established by \citet[see again the discussion in \S\ref{ss:sim-vs-assoc}]{Hill:2015cl} through the creation of SimLex-999. This inspired other similar datasets which focused on different lexical properties. For instance, SimVerb-3500 \cite{Gerz:2016emnlp} provided similarity ratings for 3,500 English verbs, whereas CARD-660 \cite{Pilehvar:2018emnlp} aimed at measuring the semantic similarity of infrequent concepts.

\vspace{1.6mm}
\noindent \textit{Semantic Similarity Datasets in Other Languages.} 
Motivated by the impact of datasets such as SimLex-999 and SimVerb-3500 on representation learning in English, a line of related work focused on creating similar resources in other languages. The dominant approach is translating and reannotating the entire original English SimLex-999 dataset, as done previously for German, Italian, and Russian \cite{Leviant:2015arxiv}, Hebrew and Croatian \cite{Mrksic:2017tacl}, and Polish \cite{Mykowiecka:2018lrec}. \namecite{Venekoski:2017nodalida} apply this process only to a subset of 300 concept pairs from the English SimLex-999. On the other hand, \citet{Camacho:2017semeval} sampled a new set of 500 English concept pairs to ensure wider topical coverage and balance across similarity spectra, and then translated those pairs to German, Italian, Spanish, and Farsi (SEMEVAL-500). A similar approach was followed by \citet{Ercan:2018coling} for Turkish, by \citet{Huang:2019arxiv} for Mandarin Chinese, and by \citet{Sakaizawa:2018lrec} for Japanese. \citet{Netisopakul:2019arxiv} translated the concatenation of SimLex-999, WordSim-353, and the English SEMEVAL-500 into Thai and then reannotated it. Finally, \citet{Barzegar:2018lrec} translated English SimLex-999 and WordSim-353 to 11 resource-rich target languages (German, French, Russian, Italian, Dutch, Chinese, Portuguese, Swedish, Spanish, Arabic, Farsi), but they did not provide details concerning the translation process and the resolution of translation disagreements. More importantly, they also did not reannotate the translated pairs in the target languages. As we discussed in \S~\ref{ss:semtyp} and reiterate later in \S\ref{s:multisimlex}, semantic differences among languages can have a profound impact on the annotation scores; particulary, we show in \S\ref{ss:typsim} that these differences even roughly define language clusters based on language affinity.

A core issue with the current datasets concerns a lack of one unified procedure that ensures the comparability of resources in different languages. 
Further, concept pairs for different languages are sourced from different corpora (e.g., direct translation of the English data versus sampling from scratch in the target language). Moreover, the previous SimLex-based multilingual datasets inherit the main deficiencies of the English original version, such as the focus on nouns and highly frequent concepts. Finally, prior work mostly focused on languages that are widely spoken and do not account for the variety of the world's languages. Our long-term goal is devising a standardized methodology to extend the coverage also to languages that are resource-lean and/or typologically diverse (e.g., Welsh, Kiswahili as in this work). 

\vspace{1.6mm}
\noindent \textit{Multilingual Datasets for Natural Language Understanding.} 
The Multi-SimLex initiative and corresponding datasets are also aligned with the recent efforts on procuring multilingual benchmarks that can help advance computational modeling of natural language understanding across different languages. For instance, pretrained multilingual language models such as multilingual \textsc{bert} \cite{devlin2018bert} or \textsc{xlm} \cite{Conneau:2019nips} are typically probed on XNLI test data \cite{Conneau:2018emnlp} for cross-lingual natural language inference. XNLI was created by translating examples from the English MultiNLI dataset, and projecting its sentence labels \cite{Williams:2018naacl}. Other recent multilingual datasets target the task of question answering based on reading comprehension: i) MLQA \cite{Lewis:2019arxiv} includes 7 languages ii) XQuAD \cite{Artetxe:2019arxiv} 10 languages; iii) TyDiQA \cite{tydiqa} 9 widely spoken typologically diverse languages. While MLQA and XQuAD result from the translation from an English dataset, TyDiQA was built independently in each language. Another multilingual dataset, PAWS-X \cite{Yang:2019emnlp}, focused on the paraphrase identification task and was created translating the original English PAWS \cite{Zhang:2019naacl} into 6 languages. We believe that Multi-SimLex can substantially contribute to this endeavor by offering a comprehensive multilingual benchmark for the fundamental lexical level relation of semantic similarity. In future work, Multi-SimLex also offers an opportunity to investigate the correlations between word-level semantic similarity and performance in downstream tasks such as QA and NLI across different languages.

\section{The Base for Multi-SimLex: Extending English SimLex-999}
\label{s:ensimlex}
In this section, we discuss the design principles behind the English (\textsc{eng}) Multi-SimLex dataset, which is the basis for all the Multi-SimLex datasets in other languages, as detailed in \S\ref{s:multisimlex}. We first argue that a new, more balanced, and more comprehensive evaluation resource for lexical semantic similarity in English is necessary. We then describe how the 1,888 word pairs contained in the \textsc{eng} Multi-SimLex were selected in such a way as to represent various linguistic phenomena within a single integrated resource. 

\vspace{1.8mm}
\noindent \textit{Construction Criteria.}
The following criteria have to be satisfied by any high-quality semantic evaluation resource, as argued by previous studies focused on the creation of such resources \cite[\textit{inter alia}]{Hill:2015cl,Gerz:2016emnlp,vulic2017hyperlex,Camacho:2017semeval}: 
\vspace{1.4mm}

\noindent \textbf{(C1) Representative and diverse.} The resource must cover the full range of diverse concepts occurring in natural language, including different word classes (e.g., nouns, verbs, adjectives, adverbs), concrete and abstract concepts, a variety of lexical fields, and different frequency ranges.

\vspace{1.4mm}
\noindent \textbf{(C2) Clearly defined.} The resource must provide a clear understanding of which semantic relation exactly is annotated and measured, possibly contrasting it with other relations. For instance, the original SimLex-999 and SimVerb-3500 explicitly focus on true semantic similarity and distinguish it from broader relatedness captured by datasets such as MEN \cite{Bruni:2014jair} or WordSim-353 \cite{Finkelstein:2002tois}. 

\vspace{1.4mm}
\noindent \textbf{(C3) Consistent and reliable.} The resource must ensure consistent annotations obtained from non-expert native speakers following simple and precise annotation guidelines.

\vspace{1.4mm}
In choosing the word pairs and constructing \textsc{eng} Multi-SimLex, we adhere to these requirements. Moreover, we follow good practices established by the research on related resources.
In particular, since the introduction of the original SimLex-999 dataset \cite{Hill:2015cl}, follow-up works have improved its construction protocol across several aspects, including: 1) coverage of more lexical fields, e.g., by relying on a diverse set of Wikipedia categories \cite{Camacho:2017semeval}, 2) infrequent/rare words \cite{Pilehvar:2018emnlp}, 3) focus on particular word classes, e.g., verbs \cite{Gerz:2016emnlp}, 4) annotation quality control \cite{Pilehvar:2018emnlp}. Our goal is to make use of these improvements towards a larger, more representative, and more reliable lexical similarity dataset in English and, consequently, in all other languages.

\vspace{1.8mm}
\noindent \textit{The Final Output: English Multi-SimLex.}
In order to ensure that the criterion C1 is satisfied, we consolidate and integrate the data already carefully sampled in prior work into a single, comprehensive, and representative dataset. This way, we can control for diversity, frequency, and other properties while avoiding to perform this time-consuming selection process from scratch. Note that, on the other hand, the word pairs chosen for English are scored from scratch as part of the entire Multi-SimLex annotation process, introduced later in \S\ref{s:multisimlex}. We now describe the external data sources for the final set of word pairs:


\vspace{1.8mm}
\noindent 1) \textit{Source: SimLex-999}. \cite{Hill:2015cl}. The English Multi-SimLex has been initially conceived as an extension of the original SimLex-999 dataset. Therefore, we include all 999 word pairs from SimLex, which span 666 noun pairs, 222 verb pairs, and 111 adjective pairs. While SimLex-999 already provides examples representing different POS classes, it does not have a sufficient coverage of different linguistic phenomena: for instance, it contains only very frequent concepts, and it does not provide a representative set of verbs \citep{Gerz:2016emnlp}.

\vspace{1.4mm}
\noindent 2) \textit{Source: SemEval-17: Task 2} \cite[henceforth SEMEVAL-500;][]{Camacho:2017semeval}. We start from the full dataset of 500 concept pairs to extract a total of 334 concept pairs for English Multi-SimLex a) which contain only single-word concepts, b) which are not named entities, c) where POS tags of the two concepts are the same, d) where both concepts occur in the top 250K most frequent word types in the English Wikipedia, and e) do not already occur in SimLex-999. The original concepts were sampled as to span all the 34 domains available as part of BabelDomains \cite{Camacho:2017eacl}, which roughly correspond to the main high-level Wikipedia categories. This ensures topical diversity in our sub-sample.

\vspace{1.4mm}
\noindent 3) \textit{Source: CARD-660} \cite{Pilehvar:2018emnlp}. 67 word pairs are taken from this dataset focused on rare word similarity, applying the same selection criteria a) to e) employed for SEMEVAL-500. Words are controlled for frequency based on their occurrence counts from the Google News data and the ukWaC corpus \cite{Baroni:2009lre}. CARD-660 contains some words that are very rare (\textit{logboat}), domain-specific (\textit{erythroleukemia}) and slang (\textit{2mrw}), which might be difficult to translate and annotate across a wide array of languages. Hence, we opt for retaining only the concept pairs above the threshold of top 250K most frequent Wikipedia concepts, as above.

\vspace{1.4mm}
\noindent 4) \textit{Source: SimVerb-3500} \cite{Gerz:2016emnlp} Since both CARD-660 and SEMEVAL-500 are heavily skewed towards noun pairs, and nouns also dominate the original SimLex-999, we also extract additional verb pairs from the verb-specific similarity dataset SimVerb-3500. We randomly sample 244 verb pairs from SimVerb-3500 that represent all similarity spectra. In particular, we add 61 verb pairs for each of the similarity intervals: $[0,1.5), [1.5,3), [3,4.5), [4.5, 6]$. Since verbs in SimVerb-3500 were originally chosen from VerbNet \cite{Kipper:2004lrec,Kipper:2008lre}, they cover a wide range of verb classes and their related linguistic phenomena.

\vspace{1.4mm}
\noindent 5) \textit{Source: University of South Florida} \cite[USF;][]{Nelson:2004usf} norms, the largest database of free association for English. In order to improve the representation of different POS classes, we sample additional adjectives and adverbs from the USF norms following the procedure established by \citet{Hill:2015cl, Gerz:2016emnlp}. This yields additional 122 adjective pairs, but only a limited number of adverb pairs (e.g., \textit{later -- never}, \textit{now -- here}, \textit{once -- twice}). Therefore, we also create a set of adverb pairs semi-automatically by sampling adjectives that can be derivationally transformed into adverbs (e.g. adding the suffix \textit{-ly}) from the USF, and assessing the correctness of such derivation in WordNet. The resulting pairs include, for instance, \textit{primarily -- mainly}, \textit{softly -- firmly}, \textit{roughly -- reliably}, etc. We include a total of 123 adverb pairs into the final English Multi-SimLex. Note that this is the first time adverbs are included into any semantic similarity dataset.
 
\vspace{1.8mm}
\noindent \textit{Fulfillment of Construction Criteria.} 
The final \textsc{eng} Multi-SimLex dataset spans 1,051 noun pairs, 469 verb pairs, 245 adjective pairs, and 123 adverb pairs.\footnote{There is a very small number of adjective and verb pairs extracted from CARD-660 and SEMEVAL-500 as well. For instance, the total number of verbs is 469 since we augment the original 222 SimLex-999 verb pairs with 244 SimVerb-3500 pairs and 3 SEMEVAL-500 pairs; and similarly for adjectives.} As mentioned above, the criterion C1 has been fulfilled by relying only on word pairs that already underwent meticulous sampling processes in prior work, integrating them into a single resource. As a consequence, Multi-SimLex allows for fine-grained analyses over different POS classes, concreteness levels, similarity spectra, frequency intervals, relation types, morphology, lexical fields, and it also includes some challenging orthographically similar examples (e.g., \textit{infection -- inflection}).\footnote{Unlike SEMEVAL-500 and CARD-660, we do not explicitly control for the equal representation of concept pairs across each similarity interval for several reasons: a) Multi-SimLex contains a substantially larger number of concept pairs, so it is possible to extract balanced samples from the full data; b) such balance, even if imposed on the English dataset, would be distorted in all other monolingual and cross-lingual datasets; c) balancing over similarity intervals arguably does not reflect a true distribution ``in the wild'' where most concepts are only loosely related or completely unrelated.} We ensure that the criteria C2 and C3 are satisfied by using similar annotation guidelines as Simlex-999, SimVerb-3500, and SEMEVAL-500 that explicitly target semantic similarity. In what follows, we outline the carefully tailored process of translating and annotating Multi-SimLex datasets in all target languages.

\section{Multi-SimLex: Translation and Annotation}
\label{s:multisimlex}
We now detail the development of the final Multi-SimLex resource, describing our language selection process, as well as translation and annotation of the resource, including the steps taken to ensure and measure the quality of this resource.  We also provide key data statistics and preliminary cross-lingual comparative analyses. 

\vspace{1.6mm}
\noindent \textit{Language Selection.}
Multi-SimLex comprises eleven languages in addition to English. The main objective for our inclusion criteria has been to balance language prominence (by number of speakers of the language) for maximum impact of the resource, while simultaneously having a diverse suite of languages based on their typological features (such as morphological type and language family). Table~\ref{tab:langs} summarizes key information about the languages currently included in Multi-SimLex. We have included a mixture of fusional, agglutinative, isolating, and introflexive languages that come from eight different language families. This includes languages that are very widely used such as Chinese Mandarin and Spanish, and low-resource languages such as Welsh and Kiswahili. We hope to further include additional languages and inspire other researchers to contribute to the effort over the lifetime of this project. 


The work on data collection can be divided into two crucial phases: 1) a translation phase where the extended English language dataset with 1,888 pairs (described in \S\ref{s:ensimlex}) is translated into eleven target languages, and 2) an annotation phase where human raters scored each pair in the translated set as well as the English set. Detailed guidelines for both phases are available online at: \url{https://multisimlex.com}.

\begin{table}[t]
\def\arraystretch{0.999}
\centering
{\small
\begin{tabularx}{\linewidth}{l YYYY}
\toprule 
{\bf Language} & {ISO 639-3} & {Family} & {Type} & {\# Speakers}\\
\cmidrule(lr){3-5}
{Chinese Mandarin} & {\sc cmn} & {Sino-Tibetan} & {Isolating} & {1.116 B} \\
{Welsh} & {\sc cym} & {IE: Celtic} & {Fusional} & {0.7 M} \\
{English} & {\sc eng} & {IE: Germanic} & {Fusional} & {1.132 B} \\
{Estonian} & {\sc est} & {Uralic} & {Agglutinative} & {1.1 M} \\
{Finnish} & {\sc fin} & {Uralic} & {Agglutinative} & {5.4 M} \\
{French} & {\sc fra} & {IE: Romance} & {Fusional} & {280 M} \\
{Hebrew} & {\sc heb} & {Afro-Asiatic} & {Introflexive} & {9 M} \\
{Polish} & {\sc pol} & {IE: Slavic} & {Fusional} & {50 M} \\
{Russian} & {\sc rus} & {IE: Slavic} & {Fusional} & {260 M} \\
{Spanish} & {\sc spa} & {IE: Romance} & {Fusional} & {534.3 M} \\
{Kiswahili} & {\sc swa} & {Niger-Congo} & {Agglutinative} & {98 M} \\
{Yue Chinese} & {\sc yue} & {Sino-Tibetan} & {Isolating} & {73.5 M} \\
\bottomrule
\end{tabularx}
}
\caption{The list of 12 languages in the Multi-SimLex multilingual suite along with their corresponding language family (IE = Indo-European), broad morphological type, and their ISO 639-3 code. The number of speakers is based on the total count of L1 and L2 speakers, according to \url{ethnologue.com}.}
\label{tab:langs}
\end{table}

\subsection{Word Pair Translation}
\label{ss:translation}

Translators for each target language were instructed to find direct or approximate translations for the 1,888 word pairs that satisfy the following rules. (1) All pairs in the translated set must be unique (i.e., no duplicate pairs); (2) Translating two words from the same English pair into the same word in the target language is not allowed (e.g., it is not allowed to translate \textit{car} and \textit{automobile} to the same Spanish word \textit{coche}). (3) The translated pairs must preserve the semantic relations between the two words when possible. This means that, when multiple translations are possible, the translation that best conveys the semantic relation between the two words found in the original English pair is selected. (4) If it is not possible to use a single-word translation in the target language, then a multi-word expression (MWE) can be used to convey the nearest possible semantics given the above points (e.g., the English word \textit{homework} is translated into the Polish MWE \textit{praca domowa}).



Satisfying the above rules when finding appropriate translations for each pair--while keeping to the spirit of the intended semantic relation in the English version--is not always straightforward. For instance, kinship terminology in Sinitic languages (Mandarin and Yue) uses different terms depending on whether the family member is older or younger, and whether the family member comes from the mother’s side or the father’s side. 
\begin{CJK*}{UTF8}{gbsn}
In Mandarin, \emph{brother} has no direct translation and can be translated as either: 哥哥 (\emph{older brother}) or 弟弟 (\emph{younger brother}). Therefore, in such cases, the translators are asked to choose the best option given the semantic context (relation) expressed by the pair in English, otherwise select one of the translations arbitrarily.  This is also used to remove duplicate pairs in the translated set, by differentiating the duplicates using a variant at each instance.
\end{CJK*}
 Further, many translation instances were resolved using near-synonymous terms in the translation.  For example, the words in the pair: \emph{wood -- timber} can only be directly translated in Estonian to \emph{puit}, and are not distinguishable. Therefore, the translators approximated the translation for \textit{timber} to the compound noun \emph{puitmaterjal} (literally: \emph{wood material}) in order to produce a valid pair in the target language. In some cases, a direct transliteration from English is used. For example, the pair: \emph{physician} and \emph{doctor} both translate to the same word in Estonian (\textit{arst}); the less formal word \emph{doktor} is used as a translation of \emph{doctor} to generate a valid pair.
 
\begin{table}[!t]
\def\arraystretch{0.9}
\centering
{\footnotesize

\begin{tabularx}{\linewidth}{rrrrrrrrrrrrr}
\toprule 
{\bf Languages:} & {\sc cmn} & {\sc cym} & {\sc est} & {\sc fin} & {\sc fra} & {\sc heb} & {\sc pol} & {\sc rus} & {\sc spa} & {\sc swa} & {\sc yue} & \textit{Avg} \\
\cmidrule{2-13}
{\bf Nouns} & {84.5} & {80.0} & {90.0} & {87.3} & {78.2} & {98.2} & {90.0} & {95.5} & {85.5} & {80.0} & {77.3} & {86.0} \\
{\bf Adjectives} & {88.5} & {88.5} & {61.5} & {73.1} & {69.2} & {100.0} & {84.6} & {100.0} & {69.2} & {88.5} & {84.6} & {82.5} \\
{\bf Verbs} & {88.0} & {74.0} & {82.0} & {76.0} & {78.0} & {100.0} & {74.0} & {100.0} & {74.0} & {76.0} & {86.0} & {82.5} \\
{\bf Adverbs} & {92.9} & {100.0} & {57.1} & {78.6} & {92.9} & {100.0} & {85.7} & {100.0} & {85.7} & {85.7} & {78.6} & {87.0} \\
{\bf Overall} & {86.5}  & {81.0}  & {82.0}  & {82.0}  & {78.0}  & {99.0}  & {85.0}  & {97.5}  & {80.5}  & {81.0}  & {80.5}  & {84.8} \\

\bottomrule
\end{tabularx}

}
\caption{Inter-translator agreement (\% of matched translated words) by independent translators using a randomly selected 100-pair English sample from the Multi-SimLex dataset, and the corresponding 100-pair samples from the other datasets.}
\label{tab:intertranslator}
\vspace{-2.5mm}
\end{table}

We measure the quality of the translated pairs by using a random sample set of 100 pairs (from the 1,888 pairs) to be translated by an independent translator for each target language. The sample is proportionally stratified according to the part-of-speech categories. The independent translator is given identical instructions to the main translator; we then measure the percentage of matched translated words between the two translations of the sample set. Table~\ref{tab:intertranslator} summarizes the inter-translator agreement results for all languages and by part-of-speech subsets. Overall across all languages, the agreement is 84.8\%, which is similar to prior work \cite{Camacho:2017semeval,Vulic:2019acl}.

\subsection{Guidelines and Word Pair Scoring}

Across all languages, 145 human annotators were asked to score all 1,888 pairs (in their given language). We finally collect at least ten valid annotations for each word pair in each language. All annotators were required to abide by the following instructions:
\vspace{1.6mm}

\noindent 1. Each annotator must assign an integer score between 0 and 6 (inclusive) indicating how semantically similar the two words in a given pair are. A score of 6 indicates very high similarity (i.e., perfect synonymy), while zero indicates no similarity.

\vspace{1.4mm}
\noindent 2. Each annotator must score the entire set of 1,888 pairs in the dataset. The pairs must not be shared between different annotators.  

\vspace{1.4mm}
\noindent 3. Annotators are able to break the workload over a period of approximately 2-3 weeks, and are able to use external sources (e.g. dictionaries, thesauri, WordNet) if required.

\vspace{1.4mm} 
\noindent 4. Annotators are kept anonymous, and are not able to communicate with each other during the annotation process.

\vspace{1.6mm}
The selection criteria for the annotators required that all annotators must be native speakers of the target language.  Preference to annotators with university education was given, but not required. Annotators were asked to complete a spreadsheet containing the translated pairs of words, as well as the part-of-speech, and a column to enter the score. The annotators did not have access to the original pairs in English.

To ensure the quality of the collected ratings, we have employed an \textit{adjudication protocol} similar to the one proposed and validated by \namecite{Pilehvar:2018emnlp}. It consists of the following three rounds:

\vspace{1.6mm}

\noindent {\bf Round 1:} All annotators are asked to follow the instructions outlined above, and to rate all 1,888 pairs with integer scores between 0 and 6.

\vspace{1.4mm}
\noindent {\bf Round 2:} We compare the scores of all annotators and identify the pairs for each annotator that have shown the most disagreement. We ask the annotators to reconsider the assigned scores for those pairs only. The annotators may chose to either change or keep the scores. As in the case with Round 1, the annotators have no access to the scores of the other annotators, and the process is anonymous.  This process gives a chance for annotators to correct 
errors or reconsider their judgments, and has been shown to be very effective in reaching 
consensus, as reported by \citet{Pilehvar:2018emnlp}.  We used a very similar procedure as \citet{Pilehvar:2018emnlp} to identify the pairs with the most disagreement; for each annotator, we marked the $i$th pair if the rated score $s_i$ falls within: $s_i \geq \mu_i + 1.5$ or $s_i \leq \mu_i - 1.5$, where $\mu_i$
is the mean of the other annotators' scores.

\vspace{1.4mm}
\noindent {\bf Round 3:} We compute the average agreement for each annotator (with the other annotators), by measuring the average Spearman's correlation against all other annotators. We discard the scores of annotators that have shown the least average agreement with all other annotators, while we maintain at least ten annotators per language by the end of this round.  The actual process is done in multiple iterations: (S1) we measure the average agreement for each annotator with every other annotator (this corresponds to the APIAA measure, see later); (S2) if we still have more than 10 valid annotators and the lowest average score is higher than in the previous iteration, we remove the lowest one, and rerun S1. Table~\ref{tab:annotators} shows the number of annotators at both the start (Round 1) and end (Round 3) of our process for each language.

\vspace{1.6mm}
We measure the agreement between annotators using two metrics, average pairwise inter-annotator agreement (APIAA), and average mean inter-annotator agreement (AMIAA). Both of these use Spearman's correlation ($\rho$) between annotators scores, the only difference is how they are averaged. They are computed as follows: 
\begin{table}[!t]
\def\arraystretch{0.99}
\centering
{\footnotesize
\begin{tabularx}{\linewidth}{l YYYYYYYYYYYY}
\toprule 
{\bf Languages:} & {\textsc{cmn}} & {\textsc{cym}} & {\textsc{eng}} & {\textsc{est}} & {\textsc{fin}} & {\textsc{fra}} & {\textsc{heb}} & {\textsc{pol}} & {\textsc{rus}} & {\textsc{spa}} & {\textsc{swa}} & {\textsc{yue}}  \\
\cmidrule(lr){2-13}
{\bf R1: Start} & {13} & {12} & {14} & {12} & {13} & {10} & {11} & {12} & {12} & {12} & {11} & {13} \\
{\bf R3: End} & {11} & {10} & {13} & {10} & {10} & {10} & {10} & {10} & {10} & {10} & {10} & {11} \\
\bottomrule
\end{tabularx}
}%
\caption{Number of human annotators. R1 = Annotation Round 1, R3 = Round 3.}
\label{tab:annotators}
\end{table}

\begin{align}
\normalfont{1) \textsc{apiaa}} = \frac{2\sum_{i,j} \rho(s_i,s_j)}{N(N-1)} \hspace{6mm} \normalfont{2) \textsc{amiaa}} = \frac{\sum_{i} \rho(s_i,\mu_{i})}{N}\,\text{, where:}\;\mu_{i} = \frac{\sum_{j, j \neq i} s_j}{N-1} 
\end{align}

where $\rho(s_i,s_j)$ is the Spearman's correlation between annotators $i$ and $j$'s scores ($s_i$,$s_j$) for all pairs in the dataset, and $N$ is the number of annotators. APIAA has been used widely as the standard measure for inter-annotator agreement, including in the original SimLex paper \cite{Hill:2015cl}. It simply averages the pairwise Spearman's correlation between all annotators. On the other hand, AMIAA compares the average Spearman's correlation of one held-out annotator with the average of all the other $N-1$ annotators, and then averages across all $N$ `held-out' annotators. It smooths individual annotator effects and arguably serves as a better upper bound than APIAA \cite[\textit{inter alia}]{Gerz:2016emnlp,vulic2017hyperlex,Pilehvar:2018emnlp}.



\begin{table}[t]
\def\arraystretch{0.999}
\centering
{\footnotesize

\begin{tabularx}{\linewidth}{lYYYYYYYYYYYY}
\toprule 
{\bf Languages:} & {\sc cmn} & {\sc cym} & {\sc eng} & {\sc est} & {\sc fin} & {\sc fra} & {\sc heb} & {\sc pol} & {\sc rus} & {\sc spa} & {\sc swa} & {\sc yue} \\
\cmidrule{2-13}
{\bf Nouns} & {0.661} & {0.622} & {0.659} & {0.558} & {0.647} & {0.698} & {0.538} & {0.606} & {0.524} & {0.582} & {0.626} & {0.727} \\
{\bf Adjectives} & {0.757} & {0.698} & {0.823} & {0.695} & {0.721} & {0.741} & {0.683} & {0.699} & {0.625} & {0.64} & {0.658} & {0.785} \\
{\bf Verbs} & {0.694} & {0.604} & {0.707} & {0.58} & {0.644} & {0.691} & {0.615} & {0.593} & {0.555} & {0.588} & {0.631} & {0.76} \\
{\bf Adverbs} & {0.699} & {0.593} & {0.695} & {0.579} & {0.646} & {0.595} & {0.561} & {0.543} & {0.535} & {0.563} & {0.562} & {0.716} \\
{\bf Overall} & {0.68}  & {0.619}  & {0.698}  & {0.583}  & {0.646}  & {0.697}  & {0.572}  & {0.609}  & {0.53}  & {0.576}  & {0.623}  & {0.733}  \\

\bottomrule
\end{tabularx}

}
\caption{Average pairwise inter-annotator agreement (APIAA). A score of $0.6$ and above indicates strong agreement.}
\label{tab:apiaa}
\end{table}

\begin{table}[t]
\def\arraystretch{0.999}
\centering
{\footnotesize
\begin{tabularx}{\linewidth}{lYYYYYYYYYYYY}
\toprule 
{\bf Languages:} & {\sc cmn} & {\sc cym} & {\sc eng} & {\sc est} & {\sc fin} & {\sc fra} & {\sc heb} & {\sc pol} & {\sc rus} & {\sc spa} & {\sc swa} & {\sc yue} \\
\cmidrule{2-13}
{\bf Nouns} & {0.757} & {0.747} & {0.766} & {0.696} & {0.766} & {0.809} & {0.68} & {0.717} & {0.657} & {0.71} & {0.725} & {0.804} \\
{\bf Adjectives} & {0.800} & {0.789} & {0.865} & {0.79} & {0.792} & {0.831} & {0.754} & {0.792} & {0.737} & {0.743} & {0.686} & {0.811} \\
{\bf Verbs} & {0.774} & {0.733} & {0.811} & {0.715} & {0.757} & {0.808} & {0.72} & {0.722} & {0.69} & {0.71} & {0.702} & {0.784} \\
{\bf Adverbs} & {0.749} & {0.693} & {0.777} & {0.697} & {0.748} & {0.729} & {0.645} & {0.655} & {0.608} & {0.671} & {0.623} & {0.716} \\
{\bf Overall} & {0.764}  & {0.742}  & {0.794}  & {0.715}  & {0.76}  & {0.812}  & {0.699}  & {0.723}  & {0.667}  & {0.703}  & {0.71}  & {0.792}  \\
\bottomrule
\end{tabularx}
}
\caption{Average mean inter-annotator agreement (AMIAA). A score of $0.6$ and above indicates strong agreement.}
\label{tab:amiaa}
\vspace{-2.5mm}
\end{table}

We present the respective APIAA and AMIAA scores in Table~\ref{tab:apiaa} and Table~\ref{tab:amiaa} for all part-of-speech subsets, as well as the agreement for the full datasets. As reported in prior work \cite{Gerz:2016emnlp,vulic2017hyperlex}, AMIAA scores are typically higher than APIAA scores. Crucially, the results indicate `strong agreement' (across all languages) using both measurements. The languages with the highest annotator agreement were French ({\sc fra}) and Yue Chinese ({\sc yue}), while Russian ({\sc rus}) had the lowest overall IAA scores. These scores, however, are still considered to be `moderately strong agreement'.

\subsection{Data Analysis}
\label{ss:data-analysis}

\begin{figure}[!t]
    \centering
    \includegraphics[width=0.85\linewidth]{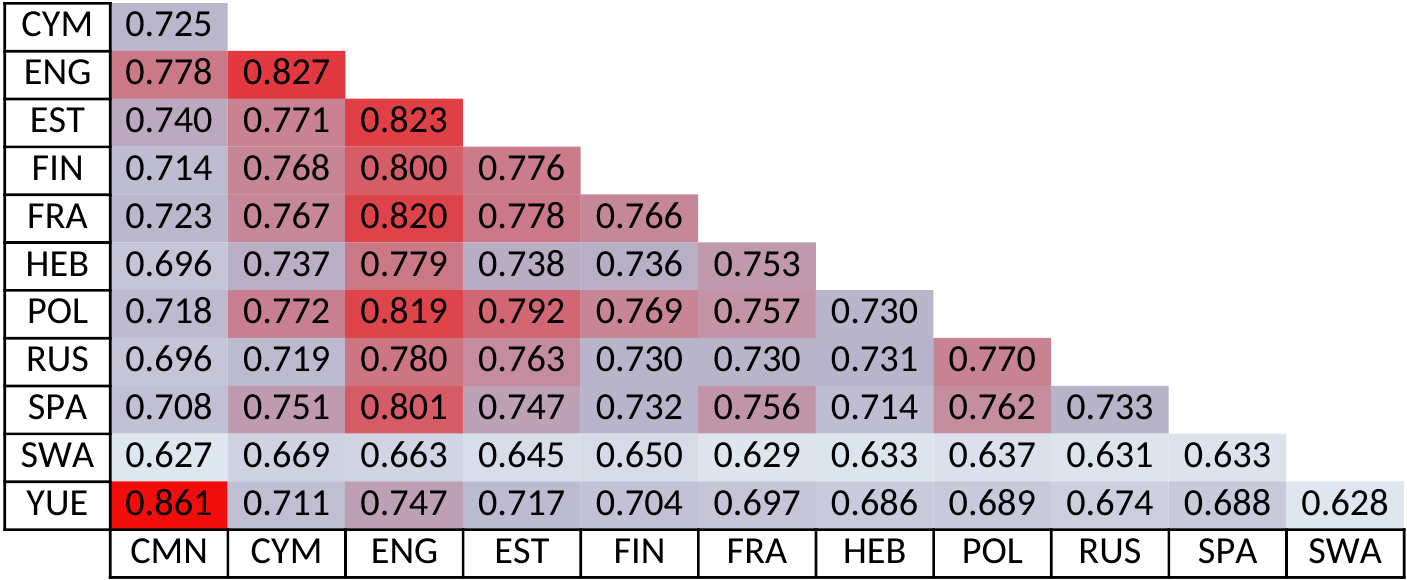}
    \caption{Spearman's correlation coefficient ($\rho$) of the similarity scores for all languages in Multi-SimLex.} 
    \label{fig:simcorr}
\end{figure}

\noindent \textit{Similarity Score Distributions.}
Across all languages, the average score (mean $=1.61$, median$=1.1$) is on the lower side of the similarity scale. However, looking closer at the scores of each language in Table~\ref{tab:mono-distrib}, we indicate notable differences in both the averages and the spread of scores. Notably, French has the highest average of similarity scores (mean$=2.61$, median$=2.5$), while Kiswahili has the lowest average (mean$=1.28$, median$=0.5$).  Russian has the lowest spread ($\sigma=1.37$), while Polish has the largest ($\sigma=1.62$). All of the languages are strongly correlated with each other, as shown in Figure~\ref{fig:simcorr}, where all of the Spearman's correlation coefficients are greater than 0.6 for all language pairs. Languages that share the same language family are highly correlated (e.g, \textsc{cmn-yue}, \textsc{rus-pol}, \textsc{est-fin}). In addition, we observe high correlations between English and most other languages, as expected. This is due to the effect of using English as the base/anchor language to create the dataset. In simple words, if one translates to two languages $L_1$ and $L_2$ starting from the same set of pairs in English, it is higly likely that $L_1$ and $L_2$ will diverge from English in different ways. Therefore, the similarity between $L_1$-\textsc{eng} and $L_2$-\textsc{eng} is expected to be higher than between $L_1$-$L_2$, especially if $L_1$ and $L_2$ are typologically dissimilar languages (e.g., \textsc{heb}-\textsc{cmn}, see Figure~\ref{fig:simcorr}). This phenomenon is well documented in related prior work \citep{Leviant:2015arxiv, Camacho:2017semeval,Mrksic:2017tacl, Vulic:2019acl}. While we acknowledge this as a slight artifact of the dataset design, it would otherwise be impossible to 
construct a semantically aligned and comprehensive dataset across a large number of languages.


\begin{table*}[!t]
 \def\arraystretch{0.99}
 \centering
 {\footnotesize
 \begin{adjustbox}{max width=\linewidth}
 \begin{tabularx}{\linewidth}{l PPPPPPPPPPPP}
 \toprule
 {\bf Lang:} & {\textsc{cmn}} & {\textsc{cym}} & {\textsc{eng}} & {\textsc{est}} & {\textsc{fin}} & {\textsc{fra}} & {\textsc{heb}} & {\textsc{pol}} & {\textsc{rus}} & {\textsc{spa}} & {\textsc{swa}} & {\textsc{yue}}  \\
 \cmidrule{2-13}
 {\bf Interval} & {} & {} & {} & {} & {} & {} & {} & {} & {} & {} & {} & {} \\
 $[0,1)$ & {56.99} & {52.01} & {50.95} & {35.01} & {47.83} & {17.69} & {28.07} & {49.36} & {50.21} & {43.96} & {61.39} & {57.89} \\
 $[1,2)$ & {8.74} & {19.54} & {17.06} & {30.67} & {21.35} & {20.39} & {35.86} & {17.32} & {22.40} & {22.35} & {11.86} & {7.84} \\
 $[2,3)$ & {13.72} & {11.97} & {12.66} & {16.21} & {12.02} & {22.03} & {16.74} & {11.86} & {11.81} & {14.83} & {9.11} & {11.76} \\
 $[3,4)$ & {11.60} & {8.32} & {8.16} & {10.22} & {10.17} & {17.64} & {8.47} & {8.95} & {8.10} & {9.38} & {7.10} & {12.98} \\
 $[4,5)$ & {6.41} & {5.83} & {6.89} & {6.25} & {5.61} & {12.55} & {6.62} & {7.57} & {5.88} & {6.78} & {6.30} & {6.89} \\
 $[5,6]$ & {2.54} & {2.33} & {4.29} & {1.64} & {2.97} & {9.64} & {4.24} & {4.93} & {1.59} & {2.70} & {4.24} & {2.65} \\
 \bottomrule
 \end{tabularx}
 \end{adjustbox}
 }
 \vspace{0mm}
 \caption{Fine-grained distribution of concept pairs over different rating intervals in each Multi-SimLex language, reported as percentages. The total number of concept pairs in each dataset is 1,888.}
 \label{tab:mono-distrib}
 \end{table*}

\begin{figure}[!t]
    \centering
    \includegraphics[width=0.85\linewidth]{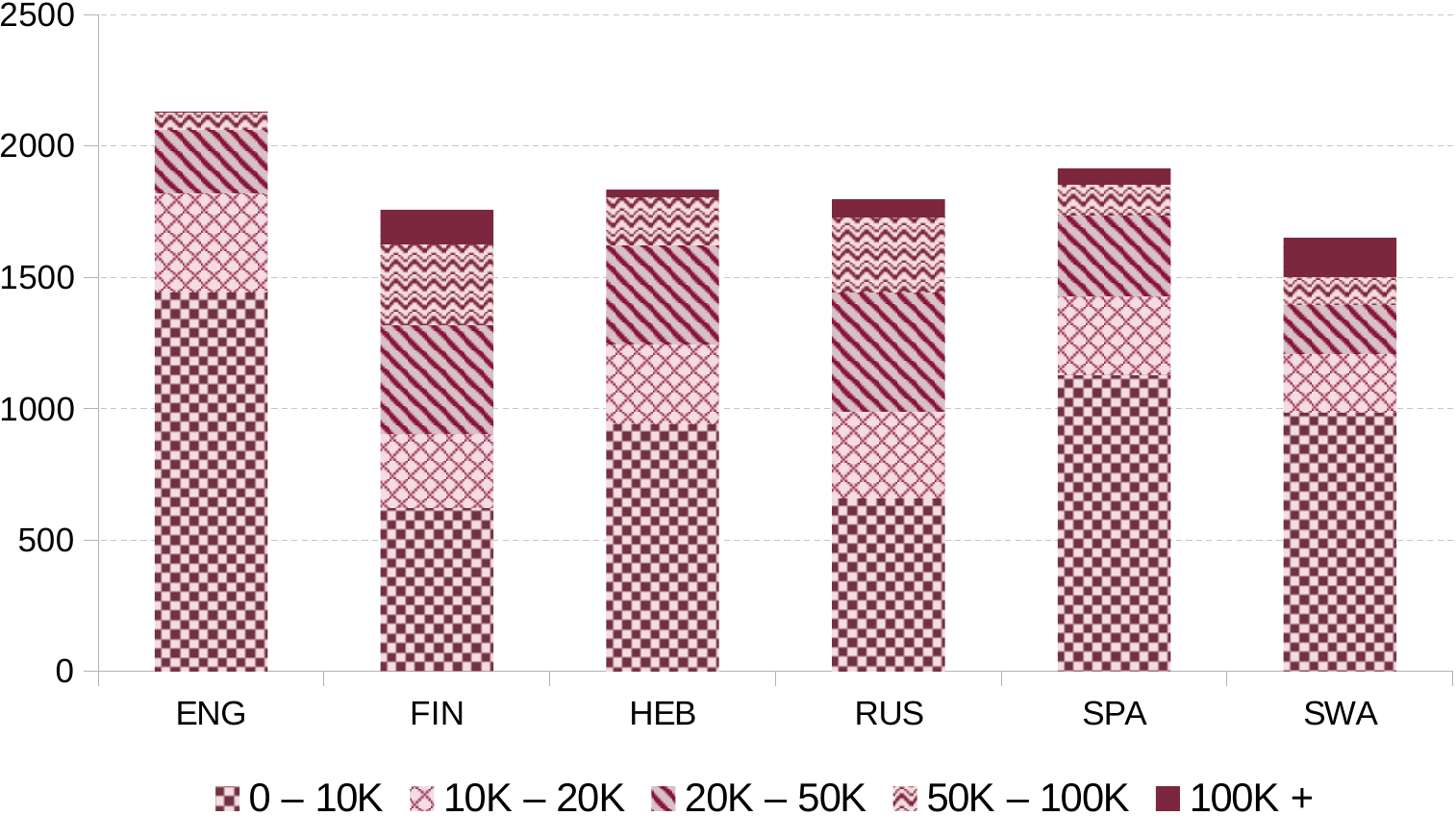}
    \caption{A distribution over different frequency ranges for words from Multi-SimLex datasets for selected languages. Multi-word expressions are excluded from the analysis.} 
    \label{fig:simfreq}
\end{figure}

We also report differences in the distribution of the frequency of words among the languages in Multi-SimLex. Figure~\ref{fig:simfreq} shows six example languages, where each bar segment shows the proportion of words in each language that occur in the given frequency range. For example, the 10K-20K segment of the bars represents the proportion of words in the dataset that occur in the list of most frequent words between the frequency rank of 10,000 and 20,000 in that language; likewise with other intervals. Frequency lists for the presented languages are derived from Wikipedia and Common Crawl corpora.\footnote{Frequency lists were obtained from fastText word vectors which are sorted by frequency: \url{https://fasttext.cc/docs/en/crawl-vectors.html}} While many concept pairs are direct or approximate translations of English pairs, we can see that the frequency distribution does vary across different languages, and is also related to inherent language properties. For instance, in Finnish and Russian, while we use infinitive forms of all verbs, conjugated verb inflections are often more frequent in raw corpora than the corresponding infinitive forms. The variance can also be partially explained by the difference in monolingual corpora size used to derive the frequency rankings in the first place: absolute vocabulary sizes are expected to fluctuate across different languages. However, it is also important to note that the datasets also contain subsets of lower-frequency and rare words, which can be used for rare word evaluations in multiple languages, in the spirit of \namecite{Pilehvar:2018emnlp}'s English rare word dataset.


\vspace{1.6mm}
\noindent \textit{Cross-Linguistic Differences.}
Table~\ref{tab:sim_examples} shows some examples of average similarity scores of English, Spanish, Kiswahili and Welsh concept pairs. Remember that the scores range from 0 to 6: the higher the score, the more similar the participants found the concepts in the pair. The examples from Table~\ref{tab:sim_examples} show evidence of both the stability of average similarity scores across languages (\emph{unlikely – friendly}, \emph{book – literature}, and \emph{vanish – disappear}), as well as language-specific differences (\emph{care – caution}). Some differences in similarity scores seem to group languages into clusters. For example, the word pair \emph{regular – average} has an average similarity score of 4.0 and 4.1 in English and Spanish, respectively, whereas in Kiswahili and Welsh the average similarity score of this pair is 0.5 and 0.8. We analyze this phenomenon in more detail in \S\ref{ss:typsim}. 

\begin{table}[!t]
\def\arraystretch{0.95}
\centering
{\footnotesize
\begin{tabularx}{\linewidth}{lYYYYY}
\toprule 
{\bf Word Pair} & {\bf POS} & {\sc eng} & {\sc spa} & {\sc swa} & {\sc cym}\\
\cmidrule{3-6}
{\bf Similar average rating}\\

{unlikely – friendly} & {ADV} & {0} & {0} & {0} & {0}\\
{book – literature} & {N} & {2.5} & {2.3} & {2.1} & {2.3}\\
{vanish – disappear} & {V} & {5.2} & {5.3} & {5.5} & {5.3}\\ \\

{\bf Different average rating}\\
{regular – average} & {ADJ} & {4} & {4.1} & {0.5} & {0.8}\\
{care – caution} & {N} & {4.1} & {5.7} & {0.2} & {3.1}\\ \\
{\bf One language higher}\\

{large – big } & {ADJ} & {5.9} & {2.7} & {3.8} & {3.8}\\
{bank – seat} & {N} & {0} & {5.1} & {0} & {0.1}\\
{sunset - evening} & {N} & {1.6} & {1.5} & {5.5} & {2.8}\\
{purely – completely} & {ADV} & {2.3} & {2.3} & {1.1} & {5.4}\\ \\

{\bf One language lower}\\
{woman – wife} & {N} & {0.9} & {2.9} & {4.1} & {4.8}\\
{amazingly – fantastically} & {ADV} & {5.1} & {0.4} & {4.1} & {4.1}\\
{wonderful – terrific} & {ADJ} & {5.3} & {5.4} & {0.9} & {5.7}\\
{promise – swear} & {V} & {4.8} & {5.3} & {4.3} & {0}\\
\bottomrule
\end{tabularx}
}
\caption{Examples of concept pairs with their similarity scores from four languages. For brevity, only the original English concept pair is included, but note that the pair is translated to all target languages, see \S\ref{ss:translation}.}
\label{tab:sim_examples}
\end{table}

There are also examples for each of the four languages having a notably higher or lower similarity score for the same concept pair than the three other languages. For example, \emph{large – big} in English has an average similarity score of 5.9, whereas Spanish, Kiswahili and Welsh speakers rate the closest concept pair in their native language to have a similarity score between 2.7 and 3.8. What is more, \emph{woman – wife} receives an average similarity of 0.9 in English, 2.9 in Spanish, and greater than 4.0 in Kiswahili and Welsh. The examples from Spanish include \emph{banco – asiento} (\emph{bank – seat}) which receives an average similarity score 5.1, while in the other three languages the similarity score for this word pair does not exceed 0.1. At the same time, the average similarity score of \emph{espantosamente – fantásticamente} (\emph{amazingly – fantastically}) is much lower in Spanish (0.4) than in other languages (4.1 – 5.1). In Kiswahili, an example of a word pair with a higher similarity score than the rest would be \emph{machweo – jioni} (\emph{sunset – evening}), having an average score of 5.5, while the other languages receive 2.8 or less, and a notably lower similarity score is given to  \emph{wa ajabu - mkubwa sana} (\emph{wonderful – terrific}), getting 0.9, while the other languages receive 5.3 or more. Welsh examples include \emph{yn llwyr - yn gyfan gwbl} (\emph{purely – completely}), which scores 5.4 among Welsh speakers but 2.3 or less in other languages, while \emph{addo – tyngu} (\emph{promise – swear}) is rated as 0 by all Welsh annotators, but in the other three languages 4.3 or more on average.

There can be several explanations for the differences in similarity scores across languages, including but not limited to cultural context, polysemy, metonymy, translation, regional and generational differences, and most commonly, the fact that words and meanings do not exactly map onto each other across languages. For example, it is likely that the other three languages do not have two separate words for describing the concepts in the concept pair: \emph{big – large}, and the translators had to opt for similar lexical items that were more distant in meaning, explaining why in English the concept pair received a much higher average similarity score than in other languages. A similar issue related to the mapping problem across languages arose in the Welsh concept pair \emph{yn llwye – yn gyfan gwbl}, where Welsh speakers agreed that the two concepts are very similar. When asked, bilingual speakers considered the two Welsh concepts more similar than English equivalents \emph{purely – completely}, potentially explaining why a higher average similarity score was reached in Welsh. The example of \emph{woman – wife} can illustrate cultural differences or another translation-related issue where the word `wife' did not exist in some languages (for example, Estonian), and therefore had to be described using other words, affecting the comparability of the similarity scores. This was also the case with the \emph{football – soccer} concept pair. The pair \emph{bank – seat} demonstrates the effect of the polysemy mismatch across languages: while `bank' has two different meanings in English, neither of them is similar to the word `seat', but in Spanish, `\emph{banco}' can mean `bank', but it can also mean `bench'. Quite naturally, Spanish speakers gave the pair \emph{banco – asiento} a higher similarity score than the speakers of languages where this polysemy did not occur. 

An example of metonymy affecting the average similarity score can be seen in the Kiswahili version of the word pair: \emph{sunset – evening} (\emph{machweo – jioni}). The average similarity score for this pair is much higher in Kiswahili, likely because the word `sunset' can act as a metonym of `evening'. The low similarity score of \emph{wonderful – terrific} in Kiswahili (\emph{wa ajabu - mkubwa sana}) can be explained by the fact that while `\emph{mkubwa sana}' can be used as `terrific' in Kiswahili, it technically means `very big', adding to the examples of translation- and mapping-related effects. The word pair \emph{amazingly – fantastically} (\emph{espantosamente – fantásticamente}) brings out another translation-related problem: the accuracy of the translation. While `\emph{espantosamente}' could arguably be translated to `amazingly', more common meanings include: `frightfully', `terrifyingly', and `shockingly', explaining why the average similarity score differs from the rest of the languages. Another problem was brought out by \emph{addo – tyngu} (\emph{promise – swear}) in Welsh, where the `\emph{tyngu}' may not have been a commonly used or even a known word choice for annotators, pointing out potential regional or generational differences in language use.

\begin{table}[!t]
\centering
\def\arraystretch{0.9}
{\footnotesize
\begin{tabularx}{\linewidth}{llYYY}
\toprule 
{\bf Language} & {\bf Word Pair} & {\bf POS} & {\bf Rating all participants agree with}\\ 
{\sc eng} & {trial – test} & {N} & {4-5}\\ 
{\sc swa} & {archbishop – bishop} & {N} & {4-5}\\ 
{\sc spa, cym} & {start – begin} & {V} & {5-6}\\ 
{\sc eng} & {smart – intelligent} & {ADJ} & {5-6}\\ 
{\sc eng, spa} & {quick – rapid} & {ADJ} & {5-6}\\ 
{\sc spa} & {circumstance – situation} & {N} & {5-6}\\ 
{\sc cym} & {football – soccer} & {N} & {5-6}\\ 
{\sc swa} & {football – soccer} & {N} & {6}\\ 
{\sc swa} & {pause – wait} & {V} & {6}\\ 
{\sc swa} & {money – cash} & {N} & {6}\\ 
{\sc cym} & {friend – buddy} & {N} & {6}\\ 
\bottomrule
\end{tabularx}
}
\caption{Examples of concept pairs with their similarity scores from four languages where all participants show strong agreement in their rating.}
\label{tab:all_agree}
\end{table}

Table~\ref{tab:all_agree} presents examples of concept pairs from English, Spanish, Kiswahili, and Welsh on which the participants agreed the most. For example, in English all participants rated the similarity of \emph{trial – test} to be 4 or 5. In Spanish and Welsh, all participants rated \emph{start – begin} to correspond to a score of 5 or 6. In Kiswahili, \emph{money – cash} received a similarity rating of 6 from every participant. While there are numerous examples of concept pairs in these languages where the participants agreed on a similarity score of 4 or higher, it is worth noting that none of these languages had a single pair where all participants agreed on either 1-2, 2-3, or 3-4 similarity rating. Interestingly, in English all pairs where all the participants agreed on a 5-6 similarity score were adjectives. 

\subsection{Effect of Language Affinity on Similarity Scores}
\label{ss:typsim}

\begin{figure}[!t]
    \centering
    \includegraphics[width=0.73\linewidth]{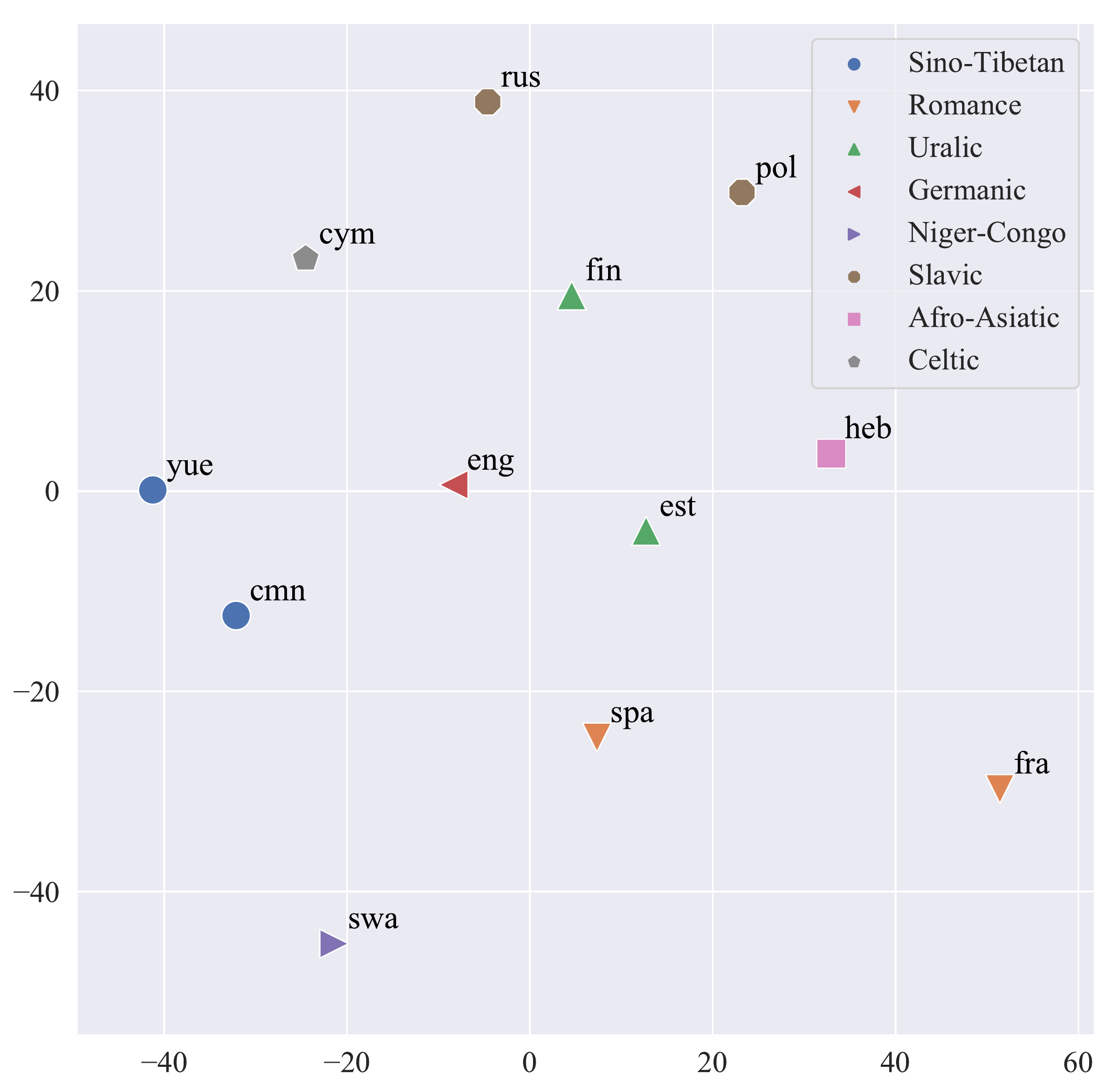}
    \caption{PCA of the language vectors resulting from the concatenation of similarity judgments for all pairs.}
    \label{fig:langvec}
\end{figure}

Based on the analysis in Figure~\ref{fig:simcorr} and inspecting the anecdotal examples in the previous section, it is evident that the correlation between similarity scores across languages is not random. To corroborate this intuition, we visualize the vectors of similarity scores for each single language by reducing their dimensionality to 2 via Principal Component Analysis \citep{pearson1901liii}. The resulting scatter plot in Figure~\ref{fig:langvec} reveals that languages from the same family or branch have similar patterns in the scores. In particular, Russian and Polish (both Slavic), Finnish and Estonian (both Uralic), Cantonese and Mandarin Chinese (both Sinitic), and Spanish and French (both Romance) are all neighbors.

In order to quantify exactly the effect of language affinity on the similarity scores, we run correlation analyses between these and language features. In particular, we extract feature vectors from URIEL \citep{littell2017uriel}, a massively multilingual typological database that collects and normalizes information compiled by grammarians and field linguists about the world's languages. In particular, we focus on information about \textit{geography} (the areas where the language speakers are concentrated), \textit{family} (the phylogenetic tree each language belongs to), and typology (including \textit{syntax}, phonological \textit{inventory}, and \textit{phonology}).\footnote{For the extraction of these features, we employed \texttt{lang2vec}: \url{github.com/antonisa/lang2vec}} Moreover, we consider typological representations of languages that are not manually crafted by experts, but rather learned from texts. \citet{malaviya2017learning} proposed to construct such representations by training language-identifying vectors end-to-end as part of neural machine translation models.

The vector for similarity judgments and the vector of linguistic features for a given language have different dimensionality. Hence, we first construct a distance matrix for each vector space, such that both columns and rows are language indices, and each cell value is the cosine distance between the vectors of the corresponding language pair. Given a set of \textit{L} languages, each resulting matrix $S$ has dimensionality of $\mathbb{R}^{|L| \times |L|}$ and is symmetrical. To estimate the correlation between the matrix for similarity judgments and each of the matrices for linguistic features, we run a Mantel test \citep{mantel1967detection}, a non-parametric statistical test based on matrix permutations that takes into account inter-dependencies among pairwise distances.

The results of the Mantel test reported in Table~\ref{fig:langvec} show that there exist statistically significant correlations between similarity judgments and geography, family, and syntax, given that $p < 0.05$ and $z > 1.96$. The correlation coefficient is particularly strong for geography ($r = 0.647$) and syntax ($r = 0.649$). The former result is intuitive, because languages in contact easily borrow and loan lexical units, and cultural interactions may result in similar cognitive categorizations. The result for syntax, instead, cannot be explained so easily, as formal properties of language do not affect lexical semantics. Instead, we conjecture that, while no causal relation is present, both syntactic features and similarity judgments might be linked to a common explanatory variable (such as geography). In fact, several syntactic properties are not uniformly spread across the globe. For instance, verbs with Verb--Object--Subject word order are mostly concentrated in Oceania \citep{wals-81}. In turn, geographical proximity leads to similar judgment patterns, as mentioned above. On the other hand, we find no correlation with phonology and inventory, as expected, nor with the bottom-up typological features from \citet{malaviya2017learning}.

\begin{table}[t]
\def\arraystretch{0.99}
\centering
\begin{tabularx}{\linewidth}{lr YYY}
\toprule
\textbf{Features} & \textbf{Dimension} & \textbf{Mantel r} & \textbf{Mantel p} & \textbf{Mantel z} \\ 
\hline
geography & 299 & 0.647 & 0.007* & 3.443 \\ 
family & 3718 & 0.329 & 0.023* & 2.711 \\ 
syntax & 103 & 0.649 & 0.007* & 3.787 \\ 
inventory & 158 & 0.155 & 0.459 & 0.782 \\ 
phonology & 28 & 0.397 & 0.046 & 1.943 \\ 
\hline
\citet{malaviya2017learning} & 512 & -0.431 & 0.264 & -1.235 \\ 
\bottomrule
\end{tabularx}

\vspace{0mm}
\caption{Mantel test on the correlation between similarity judgments from Multi-SimLex and linguistic features from typological databases.}
\label{tab:mantel}
\end{table}

\section{Cross-Lingual Multi-SimLex Datasets}
\label{s:xsimlex}
A crucial advantage of having semantically aligned monolingual datasets across different languages is the potential to create \textit{cross-lingual semantic similarity datasets}. Such datasets allow for probing the quality of cross-lingual representation learning algorithms \cite{Camacho:2017semeval,Conneau:2018iclr,Chen:2018emnlp,Doval:2018emnlp,Ruder:2019jair,Conneau:2019nips,Ruder:2019tutorial} as an intrinsic evaluation task. However, the cross-lingual datasets previous work relied upon \cite{Camacho:2017semeval} were limited to a homogeneous set of high-resource languages (e.g., English, German, Italian, Spanish) and a small number of concept pairs (all less than 1K pairs). We address both problems by 1) using a typologically more diverse language sample, and 2) relying on a substantially larger English dataset as a source for the cross-lingual datasets: 1,888 pairs in this work versus 500 pairs in the work of \namecite{Camacho:2017semeval}. As a result, each of our cross-lingual datasets contains a substantially larger number of concept pairs, as shown in Table~\ref{tab:size}. The cross-lingual {Multi-Simlex} datasets are constructed automatically, leveraging word pair translations and annotations collected in all 12 languages. This yields a total of 66 cross-lingual datasets, one for each possible combination of languages. Table~\ref{tab:size} provides the final number of concept pairs, which lie between 2,031 and 3,480 pairs for each cross-lingual dataset, whereas Table~\ref{tab:xlingex} shows some sample pairs with their corresponding similarity scores.

The automatic creation and verification of cross-lingual datasets closely follows the procedure first outlined by \namecite{Camacho:2015acl} and later adopted by \namecite{Camacho:2017semeval} (for semantic similarity) and \namecite{Vulic:2019acl} (for graded lexical entailment). First, given two languages, we intersect their aligned concept pairs obtained through translation. 
For instance, starting from the aligned pairs \textit{attroupement -- foule} in French and \textit{rahvasumm -- rahvahulk} in Estonian, we construct two cross-lingual pairs \textit{attroupement -- rahvaluk} and \textit{rahvasumm -- foule}. The scores of cross-lingual pairs are then computed as averages of the two corresponding monolingual scores. Finally, in order to filter out concept pairs whose semantic meaning was not preserved during this operation, we retain only cross-lingual pairs for which the corresponding monolingual scores $(s_s, s_t)$ differ at most by one fifth of the full scale (i.e., $\mid s_s - s_t \mid \leq 1.2$). This heuristic mitigates the noise due to cross-lingual semantic shifts \cite{Camacho:2017semeval,Vulic:2019acl}. We refer the reader to the work of \namecite{Camacho:2015acl} for a detailed technical description of the procedure. 
\begin{CJK*}{UTF8}{gbsn}
\begin{table}[!t]
\centering
\def\arraystretch{0.99}
\vspace{-0.0em}
{\scriptsize
\begin{tabularx}{\linewidth}{lllX|lllX}
\toprule
\rowcolor{Gray}
{Pair} & {Concept-1} & {Concept-2} & {Score} & {Pair} & {Concept-1} & {Concept-2} & {Score} \\
\cmidrule(lr){1-4} \cmidrule(lr){5-8}
{\textsc{cym-eng}} & {rhyddid} & {liberty} & {5.37} & {\textsc{cmn-est}} & {可能} & {optimistlikult} & {0.83} \\
{\textsc{cym-pol}} & {plentynaidd} & {niemądry} & {2.15} & {\textsc{fin-swa}} & {psykologia} & {sayansi} & {2.20} \\
{\textsc{swa-eng}} & {kutimiza} & {accomplish} & {5.24} & {\textsc{eng-fra}} & {normally} & {quotidiennement} & {2.41} \\
{\textsc{cmn-fra}} & {有弹性} & {flexible} & {4.08} & {\textsc{fin-spa}} & {auto} & {bicicleta} & {0.85} \\
{\textsc{fin-spa}} & {tietämättömyys} & {inteligencia} & {0.55} & {\textsc{cmn-yue}} & {使灰心} & {使气馁} & {4.78} \\
{\textsc{spa-fra}} & {ganador} & {candidat} & {2.15} & {\textsc{cym-swa}} & {sefyllfa} & {mazingira} & {1.90} \\
{\textsc{est-yue}} & {takso} & {巴士} & {2.08} & {\textsc{est-spa}} & {armee} & {legión} & {3.25} \\
{\textsc{eng-fin}} & {orange} & {sitrushedelmä} & {3.43} & {\textsc{fin-est}} & {halveksuva} & {põlglik} & {5.55} \\
{\textsc{spa-pol}} & {palabra} & {wskazówka} & {0.55} & {\textsc{cmn-cym}} & {学生} & {disgybl} & {4.45} \\
{\textsc{pol-swa}} & {prawdopodobnie} & {uwezekano} & {4.05} & {\textsc{pol-eng}} & {grawitacja} & {meteor} & {0.27} \\
\end{tabularx}
}
\vspace{-0.0mm}
\caption{Example concept pairs with their scores from a selection of cross-lingual Multi-SimLex datasets.}
\vspace{-0.0mm}
\label{tab:xlingex}
\end{table}
\end{CJK*}

To assess the quality of the resulting cross-lingual datasets, we have conducted a verification experiment similar to \namecite{Vulic:2019acl}.
We randomly sampled 300 concept pairs in the English-Spanish, English-French, and English-Mandarin cross-lingual datasets. Subsequently, we asked bilingual native speakers to provide similarity judgments of each pair. The Spearman's correlation score $\rho$ between automatically induced and manually collected ratings achieves $\rho \geq 0.90$ on all samples, which confirms the viability of the automatic construction procedure. 

\begin{table}[!t]
\centering
\def\arraystretch{0.99}
\vspace{-0.0em}
{\small
\begin{tabularx}{\linewidth}{l|XXXXXXXXXXXX}
& {\textsc{cmn}} & {\textsc{cym}} & {\textsc{eng}} & {\textsc{est}} & {\textsc{fin}} & {\textsc{fra}} & {\textsc{heb}} & {\textsc{pol}} & {\textsc{rus}} & {\textsc{spa}} & {\textsc{swa}} & {\textsc{yue}}  \\
\hline
{\textsc{cmn}} & \cellcolor{Gray}{1,888} & {--} & {--} & {--} & {--} & {--} & {--} & {--} & {--} & {--} & {--} & {--} \\
{\textsc{cym}} &  {3,085} & \cellcolor{Gray}{1,888} & {--} & {--} & {--} & {--} & {--} & {--} & {--} & {--} & {--} & {--} \\
{\textsc{eng}} &  {3,151} & {3,380} & \cellcolor{Gray}{1,888} & {--} & {--} & {--} & {--} & {--} & {--} & {--} & {--} & {--} \\
{\textsc{est}} &  {3,188} & {3,305} & {3,364} & \cellcolor{Gray}{1,888} & {--} & {--} & {--} & {--} & {--} & {--} & {--} & {--} \\
{\textsc{fin}} &  {3,137} & {3,274} & {3,352} & {3,386} & \cellcolor{Gray}{1,888} & {--} & {--} & {--} & {--} & {--} & {--} & {--} \\
{\textsc{fra}} &  {2,243} & {2,301} & {2,284} & {2,787} & {2,682} & \cellcolor{Gray}{1,888} & {--} & {--} & {--} & {--} & {--} & {--} \\
{\textsc{heb}} &  {3,056} & {3,209} & {3,274} & {3,358} & {3,243} & {2,903} & \cellcolor{Gray}{1,888} & {--} & {--} & {--} & {--} & {--} \\
{\textsc{pol}} &  {3,009} & {3,175} & {3,274} & {3,310} & {3,294} & {2,379} & {3,201} & \cellcolor{Gray}{1,888} & {--} & {--} & {--} & {--} \\
{\textsc{rus}} &  {3,032} & {3,196} & {3,222} & {3,339} & {3,257} & {2,219} & {3,226} & {3,209} & \cellcolor{Gray}{1,888} & {--} & {--} & {--} \\
{\textsc{spa}} &  {3,116} & {3,205} & {3,318} & {3,312} & {3,256} & {2,645} & {3,256} & {3,250} & {3,189} & \cellcolor{Gray}{1,888} & {--} & {--} \\
{\textsc{swa}} &  {2,807} & {2,926} & {2,828} & {2,845} & {2,900} & {2,031} & {2,775} & {2,819} & {2,855} & {2,811} & \cellcolor{Gray}{1,888} & {--} \\
{\textsc{yue}} &  {3,480} & {3,062} & {3,099} & {3,080} & {3,063} & {2,313} & {3,005} & {2,950} & {2,966} & {3,053} & {2,821} & \cellcolor{Gray}{1,888} \\
\end{tabularx}
}
\vspace{-0.0mm}
\caption{The sizes of all monolingual (main diagonal) and cross-lingual datasets.}
\vspace{-0.0mm}
\label{tab:size}
\end{table}

\begin{figure}[!t]
    \centering
    \begin{subfigure}[t]{0.49\textwidth}
        \centering
        \includegraphics[width=0.99\linewidth, trim={0 2mm 0 0}]{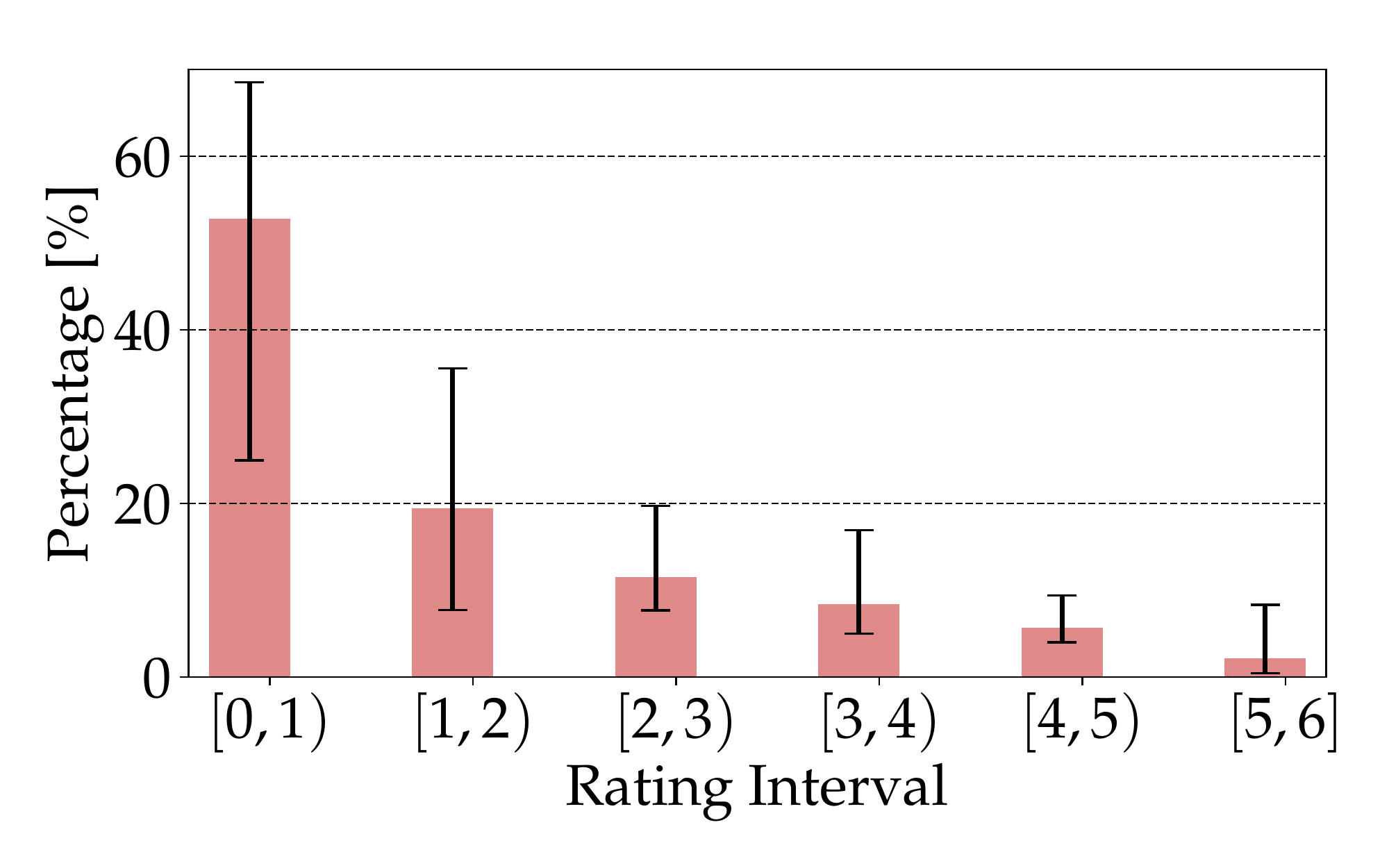}
        \caption{Rating distribution}
        \label{fig:xling-distrib}
    \end{subfigure}%
    \vspace{0mm}
    \begin{subfigure}[t]{0.49\textwidth}
        \centering
        \includegraphics[width=0.98\linewidth,trim={0 2mm 0 0}]{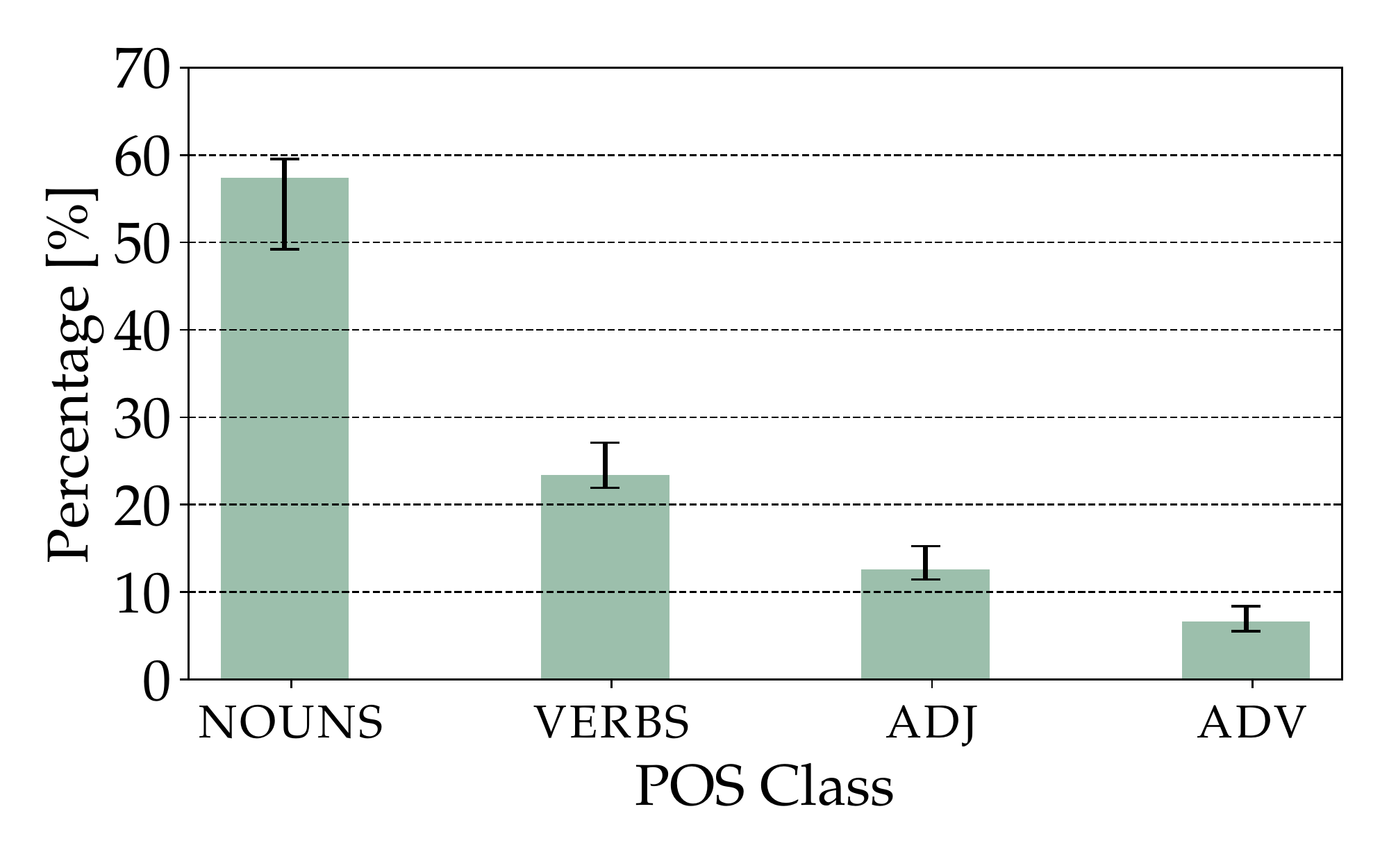}
        \caption{Distribution over POS classes}
        \label{fig:xling-pos}
    \end{subfigure}
    \vspace{0mm}
    \caption{\textbf{(a)} Rating distribution and \textbf{(b)} distribution of pairs over the four POS classes in cross-lingual Multi-SimLex datasets averaged across each of the 66 language pairs ($y$ axes plot percentages as the total number of concept pairs varies across different cross-lingual datasets). Minimum and maximum percentages for each rating interval and POS class are also plotted. 
    }
    \vspace{-0mm}
\label{fig:scores}
\end{figure}

\vspace{1.6mm}
\noindent \textit{Score and Class Distributions.}
The summary of score and class distributions across all 66 cross-lingual datasets are provided in Figure~\ref{fig:xling-distrib} and Figure~\ref{fig:xling-pos}, respectively. First, it is obvious that the distribution over the four POS classes largely adheres to that of the original monolingual Multi-SimLex datasets, and that the variance is quite low: e.g., the \textsc{eng-fra} dataset contains the lowest proportion of nouns (49.21\%) and the highest proportion of verbs (27.1\%), adjectives (15.28\%), and adverbs (8.41\%). 
On the other hand, the distribution over similarity intervals in Figure~\ref{fig:xling-distrib} shows a much greater variance. This is again expected as this pattern resurfaces in monolingual datasets (see Table~\ref{tab:mono-distrib}). It is also evident that the data are skewed towards lower-similarity concept pairs. However, due to the joint size of all cross-lingual datasets (see Table~\ref{tab:size}), even the least represented intervals contain a substantial number of concept pairs. For instance, the \textsc{rus-yue} dataset contains the least highly similar concept pairs (in the interval $[4,6]$) of all 66 cross-lingual datasets. Nonetheless, the absolute number of pairs (138) in that interval for \textsc{rus-yue} is still substantial. If needed, this makes it possible to create smaller datasets which are balanced across the similarity spectra through sub-sampling.

\section{Monolingual Evaluation of Representation Learning Models}
\label{s:monoeval}

After the numerical and qualitative analyses of the Multi-SimLex datasets provided in \S\S~\ref{ss:data-analysis}--\ref{ss:typsim}, we now benchmark a series of representation learning models on the new evaluation data. We evaluate standard static word embedding algorithms such as fastText \cite{Bojanowski:2017tacl}, as well as a range of more recent text encoders pretrained on language modeling such as multilingual BERT \citep{devlin2018bert}. These experiments provide strong baseline scores on the new Multi-SimLex datasets and offer a first large-scale analysis of pretrained encoders on word-level semantic similarity across diverse languages. In addition, the experiments now enabled by Multi-SimLex aim to answer several important questions. \textbf{(Q1)} Is it viable to extract high-quality word-level representations from  pretrained encoders receiving subword-level tokens as input? Are such representations competitive with standard static word-level embeddings? \textbf{(Q2)} What are the implications of monolingual pretraining versus (massively) multilingual pretraining for performance? \textbf{(Q3)} Do lightweight unsupervised post-processing techniques improve word representations consistently across different languages? \textbf{(Q4)} Can we effectively transfer available external lexical knowledge from resource-rich languages to resource-lean languages in order to learn word representations that distinguish between true similarity and conceptual relatedness (see the discussion in \S\ref{ss:semtyp})?

\subsection{Models in Comparison}
\label{ss:mling_models}

\textit{Static Word Embeddings in Different Languages.} First, we evaluate a standard method for inducing non-contextualized (i.e., static) word embeddings across a plethora of different languages: \textsc{fastText} (\textsc{ft}) vectors \cite{Bojanowski:2017tacl} are currently the most popular and robust choice given 1) the availability of pretrained vectors in a large number of languages \cite{Grave:2018lrec} trained on large Common Crawl (CC) plus Wikipedia (Wiki) data, and 2) their superior performance across a range of NLP tasks \cite{Mikolov:2018lrec}. In fact, \textsc{fastText} is an extension of the standard word-level CBOW and skip-gram \texttt{word2vec} models \cite{Mikolov:2013nips} that takes into account subword-level information, i.e. the contituent character n-grams of each word \cite{Zhu:2019naacl}. For this reason, \textsc{fastText} is also more suited for modeling rare words and morphologically rich languages.\footnote{We have also trained standard word-level CBOW and skip-gram with negative sampling (SGNS) on full Wikipedia dumps for several languages, but our preliminary experiments have verified that they under-perform compared to \textsc{fastText}. This finding is consistent with other recent studies demonstrating the usefulness of subword-level information \cite{Vania:2017acl,Mikolov:2018lrec,Zhu:2019naacl,Zhu:2019conll}. Therefore, we do not report the results with CBOW and SGNS for brevity.}

We rely on $300$-dimensional \textsc{ft} word vectors trained on CC+Wiki and available online for 157 languages.\footnote{\url{https://fasttext.cc/docs/en/crawl-vectors.html}} The word vectors for all languages are obtained by CBOW with position-weights, with character n-grams of length 5, a window of size 5, 10 negative examples, and 10 training epochs. 
We also probe another (older) collection of \textsc{ft} vectors, pretrained on full Wikipedia dumps of each language.\footnote{\url{https://fasttext.cc/docs/en/pretrained-vectors.html}}. The vectors are 300-dimensional, trained with the skip-gram objective for 5 epochs, with 5 negative examples, a window size set to 5, and relying on all character n-grams from length 3 to 6. 
Following prior work, we trim the vocabularies for all languages to the 200K most frequent words and compute representations for multi-word expressions by averaging the vectors of their constituent words.

\vspace{1.8mm}
\noindent \textit{Unsupervised Post-Processing.}
Further, we consider a variety of \textit{unsupervised post-processing} steps that can be applied post-training on top of any pretrained input word embedding space \textit{without} any external lexical semantic resource. So far, the usefulness of such methods has been verified only on the English language through benchmarks for lexical semantics and sentence-level tasks \cite{Mu:2018iclr}. In this paper, we assess if unsupervised post-processing is beneficial also in other languages. To this end, we apply the following post-hoc transformations on the initial word embeddings:

\vspace{1.8mm}
\noindent 1) \textit{Mean centering} (\textsc{mc}) is applied after unit length normalization to ensure that all vectors have a zero mean, and is commonly applied in data mining and analysis \cite{Bro:2003mc,Berg:2006mc}.

\vspace{1.4mm}
\noindent 2) \textit{All-but-the-top} (\textsc{abtt}) \cite{Mu:2018iclr,Tang:2019arxiv} eliminates the common mean vector and a few top dominating directions (according to PCA) from the input distributional word vectors, since they do not contribute towards distinguishing the actual semantic meaning of different words. The method contains a single (tunable) hyper-parameter $dd_{A}$ which denotes the number of the dominating directions to remove from the initial representations. Previous work has verified the usefulness of \textsc{abtt} in several English lexical semantic tasks such as semantic similarity, word analogies, and concept categorization, as well as in sentence-level text classification tasks \cite{Mu:2018iclr}.

\vspace{1.4mm}
\noindent 3) \textsc{uncovec} \cite{Artetxe:2018conll} adjusts the similarity order of an arbitrary input word embedding space, and can emphasize either syntactic or semantic information in the transformed vectors. In short, it transforms the input space $\bm{X}$ into an adjusted space $\bm{X}\bm{W}_{\alpha}$ through a linear map $\bm{W}_{\alpha}$ controlled by a single hyper-parameter $\alpha$. The $n^{\text{th}}$-order similarity transformation of the input word vector space $\bm{X}$ (for which $n=1$) can be obtained as $\bm{M}_{n}(\bm{X}) = \bm{M}_1(\bm{X}\bm{W}_{(n-1)/2})$, with $\bm{W}_{\alpha}=\bm{Q}\bm{\Gamma}^{\alpha}$, where $\bm{Q}$ and $\bm{\Gamma}$ are the matrices obtained via eigendecomposition of $\bm{X}^T\bm{X}=\bm{Q}\bm{\Gamma}\bm{Q}^T$. $\bm{\Gamma}$ is a diagonal matrix containing eigenvalues of $\bm{X}^T\bm{X}$; $\bm{Q}$ is an orthogonal matrix with eigenvectors of $\bm{X}^T\bm{X}$ as columns. While the motivation for the \textsc{uncovec} methods does originate from adjusting discrete similarity orders, note that $\alpha$ is in fact a continuous real-valued hyper-parameter which can be carefully tuned. For more technical details we refer the reader to the original work of \citet{Artetxe:2018conll}.

\vspace{1.8mm}
\noindent 
As mentioned, all post-processing methods can be seen as unsupervised retrofitting methods that, given an arbitrary input vector space $\bm{X}$, produce a perturbed/transformed output vector space $\bm{X}'$, but unlike common retrofitting methods \cite{Faruqui:2015naacl,Mrksic:2017tacl}, the perturbation is completely unsupervised (i.e., self-contained) and does not inject any external (semantic similarity oriented) knowledge into the vector space. Note that different perturbations can also be stacked: e.g., we can apply \textsc{uncovec} and then use \textsc{abtt} on top the output \textsc{uncovec} vectors. When using \textsc{uncovec} and \textsc{abtt} we always length-normalize and mean-center the data first (i.e., we apply the simple \textsc{mc} normalization). Finally, we tune the two hyper-parameters $d_A$ (for \textsc{abtt}) and $\alpha$ (\textsc{uncovec}) on the English Multi-SimLex and use the same values on the datasets of all other languages; we report results with $dd_A = 3$ or $dd_A = 10$, and $\alpha=-0.3$.

\vspace{1.8mm}
\noindent \textit{Contextualized Word Embeddings.} We also evaluate the capacity of unsupervised pretraining architectures based on language modeling objectives to reason over lexical semantic similarity. To the best of our knowledge, our article is the first study performing such analyses.  
State-of-the-art models such as \textsc{bert} \cite{devlin2018bert}, \textsc{xlm} \cite{Conneau:2019nips}, or \textsc{roberta} \cite{Liu:2019roberta} are typically very deep neural networks 
based on the Transformer architecture \cite{Vaswani:2017nips}. They receive subword-level tokens as inputs (such as WordPieces \cite{Schuster2012}) to tackle data sparsity. In output, they return contextualized embeddings, dynamic representations for words in context.

To represent words or multi-word expressions through a pretrained model, we follow prior work \cite{Liu:2019conll} and compute an input item's representation by 1) feeding it to a pretrained model \textit{in isolation}; then 2) averaging the $H$ last hidden representations for each of the item’s constituent subwords; and then finally 3) averaging the resulting subword representations to produce the final $d$-dimensional representation, where $d$ is the embedding and hidden-layer dimensionality (e.g., $d=768$ with \textsc{bert}). We opt for this approach due to its proven viability and simplicity \cite{Liu:2019conll}, as it does not require any additional corpora to condition the induction of contextualized embeddings.\footnote{We also tested another encoding method where we fed pairs instead of single words/concepts into the pretrained encoder. The rationale is that the other concept in the pair can be used as disambiguation signal. However, this method consistently led to sub-par performance across all experimental runs.} Other ways to extract the representations from pretrained models \cite{Aldarmaki:2019conll,Wu:2019arxiv,Cao:2020iclr} are beyond the scope of this work, and we will experiment with them in the future.

In other words, we treat each pretrained encoder \textsc{enc} as a black-box function to encode a single word or a multi-word expression $x$ in each language into a $d$-dimensional contextualized representation $\mathbf{x}_{\textsc{enc}} \in \mathbb{R}^d = \textsc{enc}(x)$ (e.g., $d=768$ with \textsc{bert}). As multilingual pretrained encoders, we experiment with the multilingual \textsc{bert} model (\textsc{m-bert}) \cite{devlin2018bert} and \textsc{xlm} \citep{Conneau:2019nips}. \textsc{m-bert} is pretrained on monolingual Wikipedia corpora of 102 languages (comprising all Multi-SimLex languages) with a 12-layer Transformer network, and yields $768$-dimensional representations. Since the concept pairs in Multi-SimLex are lowercased, we use the uncased version of \textsc{m-bert}.\footnote{\url{https://github.com/google-research/bert/blob/master/multilingual.md}} \textsc{m-bert} comprises all Multi-SimLex languages, and its evident ability to perform cross-lingual transfer \cite{Pires:2019acl,Wu:2019emnlp,Wang:2020iclr} also makes it a convenient baseline model for cross-lingual experiments later in \S\ref{ss:xling-eval}. 
The second multilingual model we consider, \textsc{xlm-100},\footnote{\url{https://github.com/facebookresearch/XLM}} is pretrained on Wikipedia dumps of 100 languages, and encodes each concept into a $1,280$-dimensional representation. In contrast to \textsc{m-bert}, \textsc{xlm-100} drops the next-sentence prediction objective and adds a cross-lingual masked language modeling objective.
For both encoders, the representations of each concept are computed as averages over the last $H=4$ hidden layers in all experiments, as suggested by \namecite{Wu:2019arxiv}.\footnote{In our preliminary experiments on several language pairs, we have also verified that this choice is superior to: a) using the output of only the last hidden layer (i.e., $H=1$) and b) averaging over all hidden layers (i.e., $H=12$ for the \textsc{bert-base} architecture). Likewise, using the special prepended \texttt{`[CLS]'} token rather than the constituent sub-words to encode a concept also led to much worse performance across the board.}

Besides \textsc{m-bert} and \textsc{xlm}, covering multiple languages, we also analyze the performance of ``language-specific'' \textsc{bert} and \textsc{xlm} models for the languages where they are available: Finnish, Spanish, English, Mandarin Chinese, and French. The main goal of this comparison is to study the differences in performance between multilingual ``one-size-fits-all'' encoders and language-specific encoders. For all experiments, we rely on the pretrained models released in the Transformers repository \cite{Wolf:2019hf}.\footnote{\url{github.com/huggingface/transformers}. The full list of currently supported pretrained encoders is available here: \url{huggingface.co/models}.}

Unsupervised post-processing steps devised for static word embeddings (i.e., mean-centering, \textsc{abtt}, \textsc{uncovec}) can also be applied on top of contextualized embeddings if we predefine a vocabulary of word types $V$ that will be represented in a word vector space $\mathbf{X}$. We construct such $V$ for each language as the intersection of word types covered by the corresponding CC+Wiki fastText vectors and the (single-word or multi-word) expressions appearing in the corresponding Multi-SimLex dataset. 

Finally, note that it is not feasible to evaluate a full range of available pretrained encoders within the scope of this work. Our main intention is to provide the first set of baseline results on Multi-SimLex by benchmarking a sample of most popular encoders, at the same time also investigating other important questions such as performance of static versus contextualized word embeddings, or multilingual versus language-specific pretraining. Another purpose of the experiments is to outline the wide potential and applicability of the Multi-SimLex datasets for multilingual and cross-lingual representation learning evaluation.

\begin{table*}[t]
\def\arraystretch{1.0}
\centering
{\footnotesize
\begin{adjustbox}{max width=\linewidth}
\begin{tabularx}{\linewidth}{l YYYYYYYYYYYY}
\toprule
{\bf Languages:} & {\textsc{cmn}} & {\textsc{cym}} & {\textsc{eng}} & {\textsc{est}} & {\textsc{fin}} & {\textsc{fra}} & {\textsc{heb}} & {\textsc{pol}} & {\textsc{rus}} & {\textsc{spa}} & {\textsc{swa}} & {\textsc{yue}}  \\
\cmidrule(lr){2-13}
\rowcolor{Gray}
{\bf \textsc{fastText} (CC+Wiki)}  & {\em (272)} & {\em (151)} & {\em (12)} & {\em (319)} & {\em (347)} & {\em (43)} & {\em (66)} & {\em (326)} & {\em (291)} & {\em (46)} & {\em (222)} & {\em (--)} \\
\cmidrule(lr){2-13}
{(1) \textsc{ft:init}} & {.534} & {.363} & {.528} & {.469} & {.607} & {.578} & {.450} & {.405} & {.422} & {.511} & {.439} & {--} \\
{(2) \textsc{ft:+mc}} & {.539} & {\bf .393} & {.535} & {.473} & {.621} & {.584} & {.480} & {.412} & {.424} & {.516} & {.469} & {--} \\ 
{(3) \textsc{ft:+abtt} (-3)} & {.557} & {.389} & {.536} & {\bf .495} & {.642} & {.610} & {.501} & {.427} & {.459} & {.523} & {\bf .473} & {--} \\ 
{(4) \textsc{ft:+abtt} (-10)} & {.\bf 583} & {.384} & {\bf .551} & {.476} & {.651} & {.623} & {.503} & {.455} & {\bf .500} & {\bf .542} & {.462} & {--} \\ 
{(5) \textsc{ft:+uncovec}} & {.572} & {.387} & {.550} & {.465} & {.642} & {.595} & {.501} & {.435} & {.437} & {.525} & {.437} & {--} \\
{(1)+(2)+(5)+(3)} & {.574} & {.386} & {.549} & {.476} & {\bf .655} & {.604} & {.503} & {.442} & {.452} & {.528} & {.432} & {--} \\
{(1)+(2)+(5)+(4)} & {.577} & {.376} & {.542} & {.455} & {.652} & {\bf .613} & {\bf .510} & {\bf .466} & {.491} & {.540} & {.424} & {--} \\
\hdashline
\rowcolor{Gray}
{\bf \textsc{fastText} (Wiki)}  & {\em (429)} & {\em (282)} & {\em (6)} & {\em (343)} & {\em (345)} & {\em (73)} & {\em (62)} & {\em (354)} & {\em (343)} & {\em (57)} & {\em (379)} & {\em (677)} \\
\cmidrule(lr){2-13}
{(1) \textsc{ft:init}} & {.315} & {.318} & {.436} & {.400} & {.575} & {.444} & {.428} & {.370} & {.359} & {.432} & {.332} & {.376} \\
{(2) \textsc{ft:+mc}} & {.373} & {.337} & {.445} & {\bf .404} & {.583} & {.463} & {.447} & {.383} & {.378} & {.447} & {.373} & {.427} \\
{(3) \textsc{ft:+abtt} (-3)}& {.459} & {\bf .343} & {.453} & {\bf .404} & {\bf .584} & {.487} & {.447} & {.387} & {.394} & {.456} & {\bf .423} & {\bf .429} \\
{(4) \textsc{ft:+abtt} (-10)} & {.496} & {.323} & {.460} & {.385} & {.581} & {.494} & {.460} & {\bf .401} & {\bf .400} & {\bf .477} & {.406} & {.399} \\
{(5) \textsc{ft:+uncovec}} & {.518} & {.328} & {.469} & {.375} & {.568} & {.483} & {.449} & {.389} & {.387} & {.469} & {.386} & {.394} \\
{(1)+(2)+(5)+(3)} & {\bf .526} & {.323} & {.470} & {.369} & {.564} & {\bf .495} & {.448} & {.392} & {.392} & {.473} & {.388} & {.388} \\
{(1)+(2)+(5)+(4)} & {\bf .526} & {.307} & {\bf .471} & {.355} & {.548} & {\bf .495} & {\bf .450} & {.394} & {.394} & {.476} & {.382} & {.396} \\
\hdashline
\rowcolor{Gray}
{\bf \textsc{m-bert}}  & {\em (0)} & {\em (0)} & {\em (0)} & {\em (0)} & {\em (0)} & {\em (0)} & {\em (0)} & {\em (0)} & {\em (0)} & {\em (0)} & {\em (0)} & {\em (0)} \\
\cmidrule(lr){2-13}
{(1) \textsc{m-bert:init}} & {.408} & {.033} & {.138} & {.085} & {.162} & {.115} & {.104} & {.069} & {.085} & {.145} & {.125} & {.404} \\
{(2) \textsc{m-bert:+mc}} & {.458} & {.044} & {.256} & {.122} & {.173} & {.183} & {.128} & {.097} & {.123} & {.203} & {.128} & {.469} \\
{(3) \textsc{m-bert:+abtt} (-3)} & {\bf .487} & {.056} & {.321} & {.137} & {.200} & {.287} & {.144} & {.126} & {.197} & {.299} & {.135} & {\bf .492} \\
{(4) \textsc{m-bert:+abtt} (-10)} & {.456} & {.056} & {\bf .329} & {.122} & {.164} & {\bf .306} & {.121} & {.126} & {.183} & {\bf .315} & {.136} & {.467} \\
{(5) \textsc{m-bert:+uncovec}} & {.464} & {.063} & {.317} & {\bf .144} & {\bf .213} & {.288} & {\bf .164} & {\bf .144} & {.198} & {.287} & {.143} & {.464} \\
{(1)+(2)+(5)+(3)} & {.464} & {.083} & {.326} & {.130} & {.201} & {.304} & {.149} & {.122} & {\bf .199} & {.295} & {\bf .148} & {.456} \\
{(1)+(2)+(5)+(4)} & {.444} & {\bf .086} & {.326} & {.112} & {.179} & {.305} & {.135} & {.127} & {.187} & {.285} & {.119} & {.447} \\
\bottomrule
\end{tabularx}
\end{adjustbox}
}
\vspace{0mm}
\caption{A summary of results (Spearman's $\rho$ correlation scores) on the full monolingual Multi-SimLex datasets for 12 languages. We benchmark fastText word embeddings trained on two different corpora (CC+Wiki and only Wiki) as well the multilingual \textsc{m-bert} model (see \S\ref{ss:mling_models}). Results with the initial word vectors are reported (i.e., without any unsupervised post-processing), as well as with different unsupervised post-processing methods, described in \S\ref{ss:mling_models}. The language codes are provided in Table~\ref{tab:langs}. The numbers in the parentheses (gray rows) refer to the number of OOV concepts excluded from the computation. The highest scores for each language and per model are in \textbf{bold}.}
\label{tab:mono-ft}
\end{table*}

\subsection{Results and Discussion}
\label{ss:mling-res}
The results we report are Spearman's $\rho$ coefficients of the correlation between the ranks derived from the scores of the evaluated models and the human scores provided in each Multi-SimLex dataset. The main results with static and contextualized word vectors for all test languages are summarized in Table~\ref{tab:mono-ft}. The scores reveal several interesting patterns, and also pinpoint the main challenges for future work.

\vspace{1.6mm}
\noindent \textit{State-of-the-Art Representation Models.} 
The absolute scores of CC+Wiki \textsc{ft}, Wiki \textsc{ft}, and \textsc{m-bert} are not directly comparable, because these models have different coverage. In particular, Multi-SimLex contains some out-of-vocabulary (OOV) words whose static \textsc{ft} embeddings are not available.\footnote{We acknowledge that it is possible to approximate word-level representations of OOVs with \textsc{ft} by summing the constituent n-gram embeddings as proposed by \namecite{Bojanowski:2017tacl}. However, we do not perform this step as the resulting embeddings are typically of much lower quality than non-OOV embeddings \cite{Zhu:2019naacl}.} On the other hand, \textsc{m-bert} has perfect coverage. A general comparison between CC+Wiki and Wiki \textsc{ft} vectors, however, supports the intuition that larger corpora (such as CC+Wiki) yield higher correlations. Another finding is that a single massively multilingual model such as \textsc{m-bert} cannot produce semantically rich word-level representations. Whether this actually happens because the training objective is different---or because the need to represent 100+ languages reduces its language-specific capacity---is investigated further below.

\begin{table*}[!t]
\def\arraystretch{0.99}
\centering
{\footnotesize
\begin{adjustbox}{max width=\linewidth}
\begin{tabularx}{\linewidth}{l YYYYYYYYYYYY}
\toprule
{\bf Languages:} & {\textsc{cmn}} & {\textsc{cym}} & {\textsc{eng}} & {\textsc{est}} & {\textsc{fin}} & {\textsc{fra}} & {\textsc{heb}} & {\textsc{pol}} & {\textsc{rus}} & {\textsc{spa}} & {\textsc{swa}} & {\textsc{yue}}  \\
\cmidrule{2-13}
\rowcolor{Gray}
{\bf \textsc{fastText} (CC+Wiki)} & \multicolumn{12}{c}{\textsc{ft:init}} \\
\cmidrule{2-13}
{\textsc{nouns} (1,051)} & {.561} & {.497} & {.592} & {.627} & {.709} & {.641} & {.560} & {.538} & {.526} & {.583} & {.544} & {.426} \\
{\textsc{verbs} (469)} & {.511} & {.265} & {.408} & {.379} & {.527} & {.551} & {.458} & {.384} & {.464} & {.499} & {.391} & {.252} \\
{\textsc{adj} (245)} & {.448} & {.338} & {.564} & {.401} & {.546} & {.616} & {.467} & {.284} & {.349} & {.401} & {.344} & {.288} \\
{\textsc{adv} (123)} & {.622} & {.187} & {.482} & {.378} & {.547} & {.648} & {.491} & {.266} & {.514} & {.423} & {.172} & {.103} \\
\cmidrule{2-13}
\rowcolor{Gray}
{\bf \textsc{fastText} (CC+Wiki)} & \multicolumn{12}{c}{\textsc{ft:+abtt} (-3)} \\
\cmidrule{2-13}
{\textsc{nouns}} & {.601} & {.512} & {.599} & {.621} & {.730} & {.653} & {.592} & {.585} & {.578} & {.605} & {.553} & {.431} \\
{\textsc{verbs}} & {.583} & {.305} & {.454} & {.379} & {.575} & {.602} & {.520} & {.390} & {.475} & {.526} & {.381} & {.314} \\
{\textsc{adj}} & {.526} & {.372} & {.601} & {.427} & {.592} & {.646} & {.483} & {.316} & {.409} & {.411} & {.402} & {.312} \\
{\textsc{adv}} & {.675} & {.150} & {.504} & {.397} & {.546} & {.695} & {.491} & {.230} & {.495} & {.416} & {.223} & {.081} \\
\cmidrule{2-13}
\rowcolor{Gray}
{\bf \textsc{m-bert}} & \multicolumn{12}{c}{\textsc{m-bert:+abtt} (-3)} \\
\cmidrule{2-13}
{\textsc{nouns}} & {.517} & {.091} & {.446} & {.191} & {.210} & {.364} & {.191} & {.188} & {.266} & {.418} & {.142} & {.539} \\
{\textsc{verbs}} & {.511} & {.005} & {.200} & {.039} & {.077} & {.248} & {.038} & {.107} & {.181} & {.266} & {.091} & {.503} \\
{\textsc{adj}} & {.227} & {.050} & {.226} & {.028} & {.128} & {.193} & {.044} & {.046} & {.002} & {.099} & {.192} & {.267} \\
{\textsc{adv}} & {.282} & {.012} & {.343} & {.112} & {.173} & {.390} & {.326} & {.036} & {.046} & {.207} & {161} & {.049} \\
\cmidrule{2-13}
\rowcolor{Gray}
{\bf \textsc{xlm-100}} & \multicolumn{12}{c}{\textsc{xlm:+abtt} (-3)} \\
\cmidrule{2-13}
{\textsc{all}} & {.498} & {.096} & {.270} & {.118} & {.203} & {.234} & {.195} & {.106} & {.170} & {.289} & {.130} & {.506} \\
{\textsc{nouns}} & {.551} & {.132} & {.381} & {.193} & {.238} & {.234} & {.242} & {.184} & {.292} & {.378} & {.165} & {.559} \\
{\textsc{verbs}} & {.544} & {.038} & {.169} & {.006} & {.190} & {.132} & {.136} & {.073} & {.095} & {.243} & {.047} & {.570} \\
{\textsc{adj}} & {.356} & {.140} & {.256} & {.081} & {.179} & {.185} & {.150} & {.046} & {.022} & {.100} & {.220} & {.291} \\
{\textsc{adv}} & {.284} & {.017} & {.040} & {.086} & {.043} & {.027} & {.221} & {.014} & {.022} & {.315} & {.095} & {.156} \\
\bottomrule
\end{tabularx}
\end{adjustbox}
}
\vspace{0mm}
\caption{Spearman's $\rho$ correlation scores over the four POS classes represented in Multi-SimLex datasets. 
In addition to the word vectors considered earlier in Table~\ref{tab:mono-ft}, we also report scores for another contextualized model, \textsc{xlm-100}. The numbers in parentheses refer to the total number of POS-class pairs in the original \textsc{eng} dataset and, consequently, in all other monolingual datasets. 
}
\label{tab:mono-pos}
\end{table*}

The overall results also clearly indicate that (i) there are differences in performance across different monolingual Multi-SimLex datasets, and (ii) unsupervised post-processing is universally useful, and can lead to huge improvements in correlation scores for many languages. In what follows, we also delve deeper into these analyses.

\vspace{1.6mm}
\noindent \textit{Impact of Unsupervised Post-Processing.} First, the results in Table~\ref{tab:mono-ft} suggest that applying dimension-wise mean centering to the initial vector spaces has positive impact on word similarity scores in all test languages and for all models, both static and contextualized (see the \textsc{+mc} rows in Table~\ref{tab:mono-ft}). \citet{Mimno:2017emnlp} show that distributional word vectors have a tendency towards narrow clusters in the vector space (i.e., they occupy a narrow cone in the vector space and are therefore anisotropic \cite{Mu:2018iclr,Ethay:2019emnlp}), and are prone to the undesired effect of hubness \cite{Radovanovic:2010jmlr,Lazaridou:2015acl}.\footnote{Hubness can be defined as the tendency of some points/vectors (i.e., ``hubs'') to be nearest neighbors of many points in a high-dimensional (vector) space \cite{Radovanovic:2010jmlr,Lazaridou:2015acl,Conneau:2018iclr}} Applying dimension-wise mean centering has the effect of spreading the vectors across the hyper-plane and mitigating the hubness issue, which consequently improves word-level similarity, as it emerges from the reported results. Previous work has already validated the importance of mean centering for clustering-based tasks \cite{Suzuki:2013emnlp}, bilingual lexicon induction with cross-lingual word embeddings \cite{Artetxe:2018aaai,Zhang:2019acl,Vulic:2019we}, and for modeling lexical semantic change \cite{Schlechtweg:2019acl}. However, to the best of our knowledge, the results summarized in Table~\ref{tab:mono-ft} are the first evidence that also confirms its importance for semantic similarity in a wide array of languages. In sum, as a general rule of thumb, we suggest to always mean-center representations for semantic tasks.

The results further indicate that additional post-processing methods such as \textsc{abtt} and \textsc{uncovec} on top of mean-centered vector spaces can lead to further gains in most languages. The gains are even visible for languages which start from high correlation scores: for instance., \textsc{cmn} with CC+Wiki \textsc{ft} increases from 0.534 to 0.583, from 0.315 to 0.526 with Wiki \textsc{ft}, and from 0.408 to 0.487 with \textsc{m-bert}. Similarly, for \textsc{rus} with CC+Wiki \textsc{ft} we can improve from 0.422 to 0.500, and for \textsc{fra} the scores improve from 0.578 to 0.613. There are additional similar cases reported in Table~\ref{tab:mono-ft}. 

Overall, the unsupervised post-processing techniques seem universally useful across languages, but their efficacy and relative performance does vary across different languages. Note that we have not carefully fine-tuned the hyper-parameters of the evaluated post-processing methods, so additional small improvements can be expected for some languages. The main finding, however, is that these post-processing techniques are robust to semantic similarity computations  beyond English, and are truly language independent. For instance, removing dominant latent (PCA-based) components from word vectors emphasizes semantic differences between different concepts, as only shared non-informative latent semantic knowledge is removed from the representations. 

In summary, pretrained word embeddings do contain more information pertaining to semantic similarity than revealed in the initial vectors. This way, we have corroborated the hypotheses from prior work \cite{Mu:2018iclr,Artetxe:2018conll} which were not previously empirically verified on other languages due to a shortage of evaluation data; this gap has now been filled with the introduction of the Multi-SimLex datasets. In all follow-up experiments, we always explicitly denote which post-processing configuration is used in evaluation.

\vspace{1.6mm}
\noindent \textit{POS-Specific Subsets.} We present the results for subsets of word pairs grouped by POS class in Table~\ref{tab:mono-pos}. Prior work based on English data showed that representations for nouns are typically of higher quality than those for the other POS classes \cite{Schwartz:2015conll,Schwartz:2016naacl,Vulic:2017conll}. We observe a similar trend in other languages as well. This pattern is consistent across different representation models and can be attributed to several reasons. First, verb representations need to express a rich range of syntactic and semantic behaviors rather than purely referential features \cite{Gruber:1976book,Levin:1993book,Kipper:2008lre}. Second, low correlation scores on the adjective and adverb subsets in some languages (e.g., \textsc{pol}, \textsc{cym}, \textsc{swa}) might be due to their low frequency in monolingual texts, which yields unreliable representations. In general, the variance in performance across different word classes warrants further research in class-specific representation learning \cite{Baker:2014emnlp,Vulic:2017conll}. The scores further attest the usefulness of unsupervised post-processing as almost all class-specific correlation scores are improved by applying mean-centering and \textsc{abtt}. Finally, the results for \textsc{m-bert} and \textsc{xlm-100} in Table~\ref{tab:mono-pos} further confirm that massively multilingual pretraining cannot yield reasonable semantic representations for many languages: in fact, for some classes they display no correlation with human ratings at all. 

\vspace{1.6mm}
\noindent \textit{Differences across Languages.}
Naturally, the results from Tables~\ref{tab:mono-ft} and \ref{tab:mono-pos} also reveal that there is variation in performance of both static word embeddings and pretrained encoders across different languages. 
Among other causes, the lowest absolute scores with \textsc{ft} are reported for languages with least resources available to train monolingual word embeddings, such as Kiswahili, Welsh, and Estonian. The low performance on Welsh is especially indicative: Figure~\ref{fig:simcorr} shows that the ratings in the Welsh dataset match up very well with the English ratings, but we cannot achieve the same level of correlation in Welsh with Welsh \textsc{ft} word embeddings. Difference in performance between two closely related languages, \textsc{est} (low-resource) and \textsc{fin} (high-resource), provides additional evidence in this respect.

The highest reported scores with \textsc{m-bert} and \textsc{xlm-100} are obtained for Mandarin Chinese and Yue Chinese: this effectively points to the weaknesses of massively multilingual training with a joint subword vocabulary spanning 102 and 100 languages. Due to the difference in scripts, ``language-specific'' subwords for \textsc{yue} and \textsc{cmn} do not need to be shared across a vast amount of languages and the quality of their representation remains unscathed. This effectively means that \textsc{m-bert}'s subword vocabulary contains plenty of \textsc{cmn}-specific and \textsc{yue}-specific subwords which are exploited by the encoder when producing \textsc{m-bert}-based representations. Simultaneously, higher scores with \textsc{m-bert} (and \textsc{xlm} in Table~\ref{tab:mono-pos}) are reported for resource-rich languages such as French, Spanish, and English, which are better represented in \textsc{m-bert}'s training data. We also observe lower absolute scores (and a larger number of OOVs) for languages with very rich and productive morphological systems such as the two Slavic languages (Polish and Russian) and Finnish. Since Polish and Russian are known to have large Wikipedias and Common Crawl data \cite{Conneau:2019arxiv} (e.g., their Wikipedias are in the top 10 largest Wikipedias worldwide), the problem with coverage can be attributed exactly to the proliferation of morphological forms in those languages.



Finally, while Table~\ref{tab:mono-ft} does reveal that unsupervised post-processing is useful for all languages, it also demonstrates that peak scores are achieved with different post-processing configurations. This finding suggests that a more careful language-specific fine-tuning is indeed needed to refine word embeddings towards semantic similarity. We plan to inspect the relationship between post-processing techniques and linguistic properties in more depth in future work.

\vspace{1.8mm}
\noindent \textit{Multilingual vs. Language-Specific Contextualized Embeddings.} Recent work has shown that---despite the usefulness of massively multilingual models such as \textsc{m-bert} and \textsc{xlm-100} for zero-shot cross-lingual transfer \cite{Pires:2019acl,Wu:2019emnlp}---stronger results in downstream tasks for a particular language can be achieved by pretraining language-specific models on language-specific data. 

In this experiment, motivated by the low results of \textsc{m-bert} and \textsc{xlm-100} (see again Table~\ref{tab:mono-pos}), we assess if monolingual pretrained encoders can produce higher-quality word-level representations than multilingual models. Therefore, we evaluate language-specific \textsc{bert} and \textsc{xlm} models for a subset of the Multi-SimLex languages for which such models are currently available: Finnish \cite{Virtanen:2019arxiv} (\textsc{bert-base} architecture, uncased), French \cite{Le:2019flaubert} (the FlauBERT model based on \textsc{xlm}), English (\textsc{bert-base}, uncased), Mandarin Chinese (\textsc{bert-base}) \cite{devlin2018bert} and Spanish (\textsc{bert-base}, uncased). In addition, we also evaluate a series of pretrained encoders available for English: (i) \textsc{bert-base}, \textsc{bert-large}, and \textsc{bert-large} with whole word masking (\textsc{wwm}) from the original work on BERT \cite{devlin2018bert}, (ii) monolingual ``English-specific'' \textsc{xlm} \cite{Conneau:2019nips}, and (iii) two models which employ parameter reduction techniques to build more compact encoders: \textsc{albert-b} uses a configuration similar to \textsc{bert-base}, while \textsc{albert-l} is similar to \textsc{bert-large}, but with an $18\times$ reduction in the number of parameters \cite{Lan:2020iclr}.\footnote{All models and their further specifications are available at the following link: \url{https://huggingface.co/models}.}

\begin{figure}[!t]
    \centering
    \begin{subfigure}[t]{0.49\textwidth}
        \centering
        \includegraphics[width=0.99\linewidth, trim={0 2mm 0 0}]{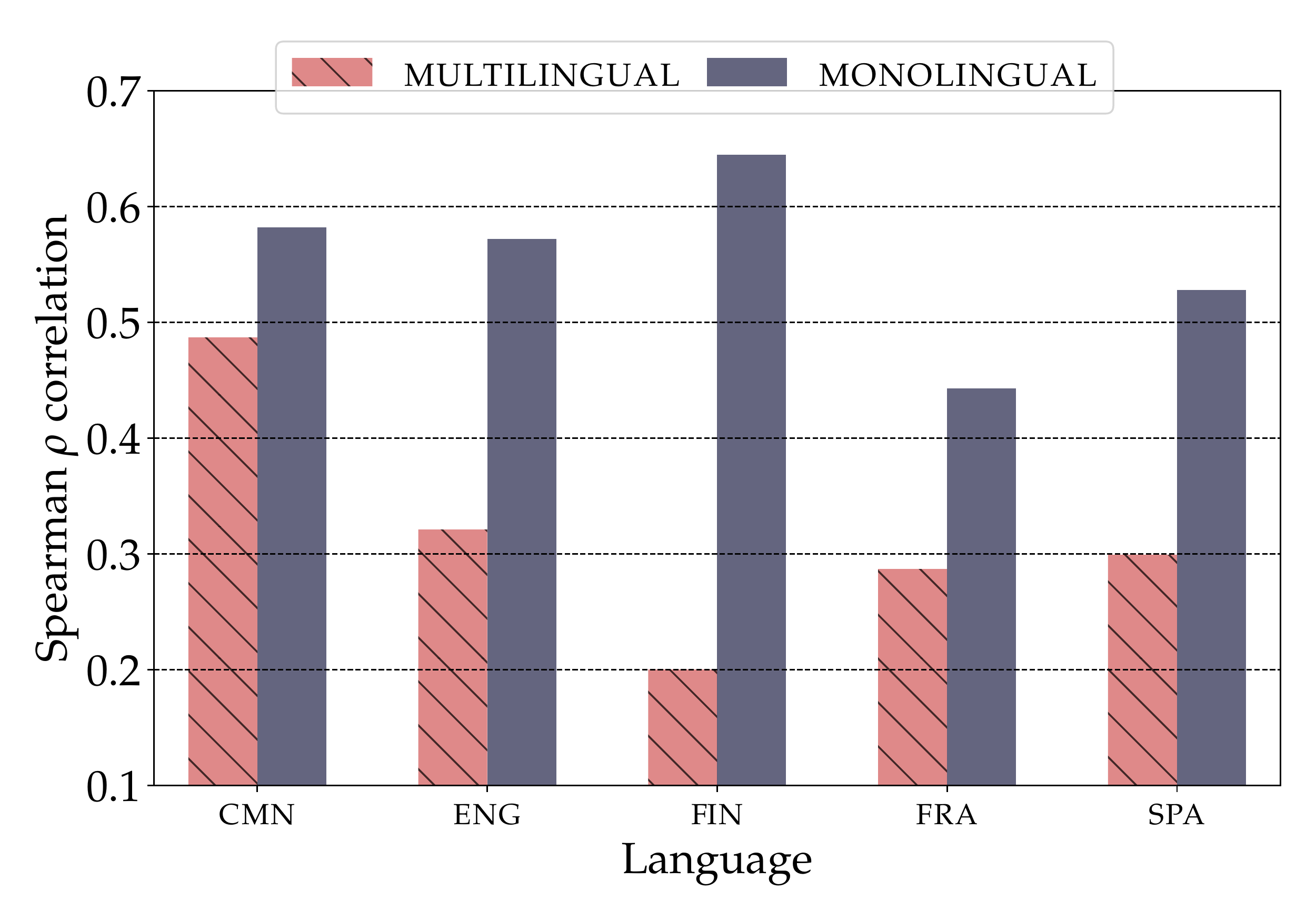}
        \caption{Monolingual vs multilingual}
        \label{fig:mono-multi}
    \end{subfigure}%
    \vspace{0mm}
    \begin{subfigure}[t]{0.49\textwidth}
        \centering
        \includegraphics[width=0.98\linewidth,trim={0 2mm 0 0}]{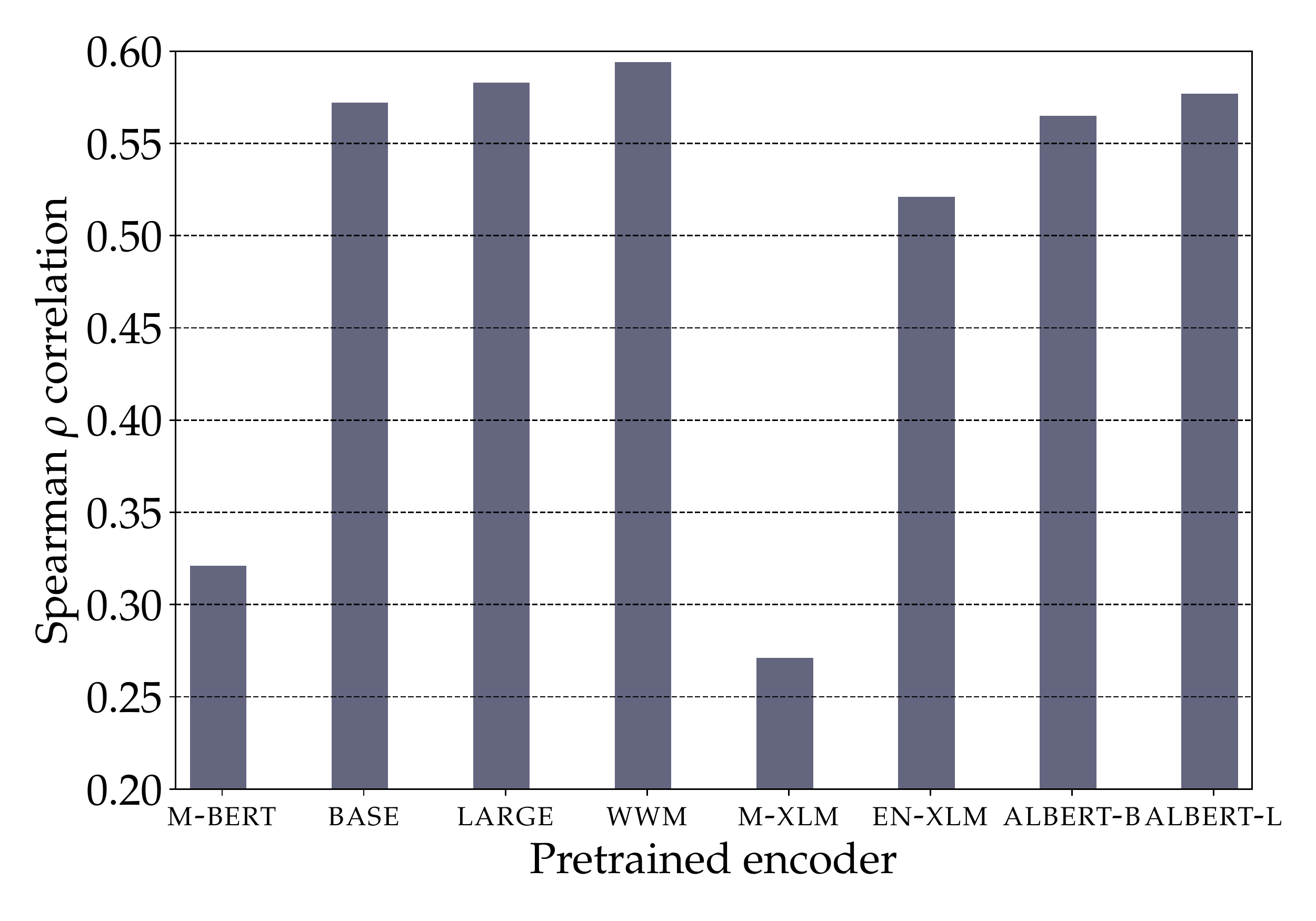}
        \caption{Pretrained \textsc{eng} encoders}
        \label{fig:eng-enc}
    \end{subfigure}
    \vspace{0mm}
    \caption{\textbf{(a)} A performance comparison between monolingual pretrained language encoders and massively multilingual encoders. For four languages (\textsc{cmn}, \textsc{eng}, \textsc{fin}, \textsc{spa}), we report the scores with monolingual uncased \textsc{bert-base} architectures and multilingual uncased \textsc{m-bert} model, while for \textsc{fra} we report the results of the multilingual \textsc{xlm-100} architecture and a monolingual French FlauBERT model \cite{Le:2019flaubert}, which is based on the same architecture as \textsc{xlm-100}. \textbf{(b)} A comparison of various pretrained encoders available for English. All these models are post-processed via \textsc{abtt} (-3).}
    \vspace{-0mm}
\label{fig:encoders}
\end{figure}

From the results in Table~\ref{fig:encoders}, it is clear that monolingual pretrained encoders yield much more reliable word-level representations. The gains are visible even for languages such as \textsc{cmn} which showed reasonable performance with \textsc{m-bert} and are substantial on all test languages. This further confirms the validity of language-specific pretraining in lieu of multilingual training, if sufficient monolingual data are available. 
Moreover, a comparison of pretrained English encoders in Figure~\ref{fig:eng-enc} largely follows the intuition: the larger \textsc{bert-large} model yields slight improvements over \textsc{bert-base}, and we can improve a bit more by relying on word-level (i.e., lexical-level) masking.
Finally, light-weight \textsc{albert} model variants are quite competitive with the original \textsc{bert} models, with only modest drops reported, and \textsc{albert-l} again outperforms \textsc{albert-b}. Overall, it is interesting to note that the scores obtained with monolingual pretrained encoders are on a par with or even outperform static \textsc{ft} word embeddings: this is a very intriguing finding per se as it shows that such subword-level models trained on large corpora can implicitly capture rich lexical semantic knowledge.

\vspace{1.8mm}
\noindent \textit{Similarity-Specialized Word Embeddings.} 
Conflating distinct lexico-semantic relations is a well-known property of distributional representations \cite{Turney:2010jair,Melamud:2016naacl}. Semantic specialization fine-tunes distributional spaces to emphasize a particular lexico-semantic relation in the transformed space by injecting external lexical knowledge \cite{Glavas:2019tutorial}. Explicitly discerning between true semantic similarity (as captured in Multi-SimLex) and broad conceptual relatedness benefits a number of tasks, as discussed in \S\ref{ss:sim-vs-assoc}.\footnote{For an overview of specialization methods for semantic similarity, we refer the interested reader to the recent tutorial \cite{Glavas:2019tutorial}.} 
Since most languages lack dedicated lexical resources, however, one viable strategy to steer monolingual word vector spaces to emphasize semantic similarity is through cross-lingual transfer of lexical knowledge, usually through a shared cross-lingual word vector space \cite{Ruder:2019jair}. Therefore, we evaluate the effectiveness of specialization transfer methods using Multi-SimLex as our multilingual test bed.

We evaluate a current state-of-the-art cross-lingual specialization transfer method with minimal requirements, put forth recently by \namecite{Ponti:2019emnlp}.\footnote{We have also evaluated other specialization transfer methods, e.g., \cite{Glavas:2018acl,Ponti:2018emnlp}, but they are consistently outperformed by the method of \namecite{Ponti:2019emnlp}.} In a nutshell, their \textsc{li-postspec} method is a multi-step procedure that operates as follows. First, the knowledge about semantic similarity is extracted from WordNet in the form of triplets, that is, linguistic constraints $(w_1, w_2, r)$, where $w_1$ and $w_2$ are two concepts, and $r$ is a relation between them obtained from WordNet (e.g., synonymy or antonymy). The goal is to ``attract'' synonyms closer to each other in the transformed vector space as they reflect true semantic similarity, and ``repel'' antonyms further apart. In the second step, the linguistic constraints are translated from English to the target language via a shared cross-lingual word vector space. To this end, following \namecite{Ponti:2019emnlp} we rely on cross-lingual word embeddings (CLWEs) \cite{Joulin:2018emnlp} available online, which are based on Wiki \textsc{ft} vectors.\footnote{\url{https://fasttext.cc/docs/en/aligned-vectors.html}; for target languages for which there are no pretrained CLWEs, we induce them following the same procedure of \namecite{Joulin:2018emnlp}.} Following that, a constraint refinement step is applied in the target language which aims to eliminate the noise inserted during the translation process. This is done by training a relation classification tool: it is trained again on the English linguistic constraints and then used on the translated target language constraints, where the transfer is again enabled via a shared cross-lingual word vector space.\footnote{We again follow \namecite{Ponti:2019emnlp} and use a state-of-the-art relation classifier \cite{Glavas:2018naacl}. We refer the reader to the original work for additional technical details related to the classifier design.} Finally, a state-of-the-art monolingual specialization procedure from \namecite{Ponti:2018emnlp} injects the (now target language) linguistic constraints into the target language distributional space.

The scores are summarized in Table~\ref{tab:mono-special}. Semantic specialization with \textsc{li-postspec} leads to substantial improvements in correlation scores for the majority of the target languages, demonstrating the importance of external semantic similarity knowledge for semantic similarity reasoning. However, we also observe deteriorated performance for the three target languages which can be considered the lowest-resource ones in our set: \textsc{cym}, \textsc{swa}, \textsc{yue}. We hypothesize that this occurs due to the inferior quality of the underlying monolingual Wikipedia word embeddings, which generates a chain of error accumulations. In particular, poor distributional word estimates compromise the alignment of the embedding spaces, which in turn results in increased translation noise, and reduced refinement ability of the relation classifier. On a high level, this ``poor get poorer'' observation again points to the fact that one of the primary causes of low performance of resource-low languages in semantic tasks is the sheer lack of even unlabeled data for distributional training. On the other hand, as we see from Table~\ref{tab:mono-pos}, typological dissimilarity between the source and the target does not deteriorate the effectiveness of semantic specialization. In fact, \textsc{li-postspec} does yield substantial gains also for the typologically distant targets such as \textsc{heb}, \textsc{cmn}, and \textsc{est}. The critical problem indeed seems to be insufficient raw data for monolingual distributional training.


\begin{table*}[!t]
\def\arraystretch{0.99}
\centering
{\footnotesize
\begin{adjustbox}{max width=\linewidth}
\begin{tabularx}{\linewidth}{l YYYYYYYYYYYY}
\toprule
{\bf Languages:} & {\textsc{cmn}} & {\textsc{cym}} & {\textsc{eng}} & {\textsc{est}} & {\textsc{fin}} & {\textsc{fra}} & {\textsc{heb}} & {\textsc{pol}} & {\textsc{rus}} & {\textsc{spa}} & {\textsc{swa}} & {\textsc{yue}}  \\
\cmidrule{2-13}
\rowcolor{Gray}
{\bf \textsc{fastText} (Wiki)}  & {\em (429)} & {\em (282)} & {\em (6)} & {\em (343)} & {\em (345)} & {\em (73)} & {\em (62)} & {\em (354)} & {\em (343)} & {\em (57)} & {\em (379)} & {\em (677)} \\
\cmidrule{2-13}
\cmidrule(lr){2-13}
{\textsc{ft:init}} & {.315} & {\bf .318} & {--} & {.400} & {.575} & {.444} & {.428} & {.370} & {.359} & {.432} & {\bf .332} & {\bf .376} \\
{\textsc{li-postspec}} & \textbf{.584} & {.204} & {--} & \textbf{.515} & \textbf{.619} & \textbf{.601} & \textbf{.510} & \textbf{.531} & \textbf{.547} & \textbf{.635} & {.238} & {.267} \\
\bottomrule
\end{tabularx}
\end{adjustbox}
}
\vspace{0mm}
\caption{The impact of vector space specialization for semantic similarity. The scores are reported using the current state-of-the-art specialization transfer \textsc{li-postspec} method of \namecite{Ponti:2019emnlp}, relying on English as a resource-rich source language and the external lexical semantic knowledge from the English WordNet.}
\label{tab:mono-special}
\end{table*}

\section{Cross-Lingual Evaluation}
\label{ss:xling-eval}
Similar to monolingual evaluation in \S\ref{s:monoeval}, we now evaluate several state-of-the-art cross-lingual representation models on the suite of 66 automatically constructed cross-lingual Multi-SimLex datasets. Again, note that evaluating a full range of cross-lingual models available in the rich prior work on cross-lingual representation learning is well beyond the scope of this article. We therefore focus our cross-lingual analyses on several well-established and indicative state-of-the-art cross-lingual models, again spanning both static and contextualized cross-lingual word embeddings.

\subsection{Models in Comparison}
\label{ss:xling-models}
\noindent \textit{Static Word Embeddings.} We rely on a state-of-the-art mapping-based method for the induction of cross-lingual word embeddings (CLWEs): \textsc{vecmap} \cite{Artetxe:2018acl}. The core idea behind such mapping-based or projection-based approaches is to learn a post-hoc alignment of independently trained monolingual word embeddings \cite{Ruder:2019jair}. Such methods have gained popularity due to their conceptual simplicity and competitive performance coupled with reduced bilingual supervision requirements: they support CLWE induction with only as much as a few thousand word translation pairs as the bilingual supervision \cite{Mikolov:2013arxiv,Xing:2015naacl,Upadhyay:2016acl,Ruder:2019tutorial}. More recent work has shown that CLWEs can be induced with even weaker supervision from small dictionaries spanning several hundred pairs \cite{Vulic:2016acl,Vulic:2019we}, identical strings \cite{Smith:2017iclr}, or even only shared numerals \cite{Artetxe:2017acl}. In the extreme, \textit{fully unsupervised} projection-based CLWEs extract such seed bilingual lexicons from scratch on the basis of monolingual data only \cite[\textit{inter alia}]{Conneau:2018iclr,Artetxe:2018acl,Hoshen:2018emnlp,Alvarez:2018emnlp,Chen:2018emnlp,Mohiuddin:2019naacl}.

Recent empirical studies \cite{Glavas:2019acl,Vulic:2019we,doval2019onthe} have compared a variety of unsupervised and weakly supervised mapping-based CLWE methods, and \textsc{vecmap} emerged as the most robust and very competitive choice. Therefore, we focus on 1) its fully unsupervised variant (\textsc{unsuper}) in our comparisons. For several language pairs, we also report scores with two other \textsc{vecmap} model variants: 2) a supervised variant which learns a mapping based on an available seed lexicon (\textsc{super}), and 3) a supervised variant \textit{with self-learning} (\textsc{super+sl}) which iteratively increases the seed lexicon and improves the mapping gradually. For a detailed description of these variants, we refer the reader to recent work \cite{Artetxe:2018acl,Vulic:2019we}. We again use CC+Wiki \textsc{ft} vectors as initial monolingual word vectors, except for \textsc{yue} where Wiki \textsc{ft} is used. The seed dictionaries of two different sizes (1k and 5k translation pairs) are based on PanLex \cite{Kamholz:2014lrec}, and are taken directly from prior work \cite{Vulic:2019we},\footnote{\url{https://github.com/cambridgeltl/panlex-bli}} or extracted from PanLex following the same procedure as in the prior work. 

\vspace{1.6mm}
\noindent \textit{Contextualized Cross-Lingual Word Embeddings.} We again evaluate the capacity of (massively) multilingual pretrained language models, \textsc{m-bert} and \textsc{xlm-100}, to reason over cross-lingual lexical similarity. Implicitly, such an evaluation also evaluates ``the intrinsic quality'' of shared cross-lingual word-level vector spaces induced by these methods, and their ability to boost cross-lingual transfer between different language pairs. We rely on the same procedure of aggregating the models' subword-level parameters into word-level representations, already described in \S\ref{ss:mling_models}.

As in monolingual settings, we can apply unsupervised post-processing steps such as \textsc{abtt} to both static and contextualized cross-lingual word embeddings.

\begin{table}[!t]
\centering
\def\arraystretch{0.99}
\vspace{-0.0em}
{\small
\begin{tabularx}{\linewidth}{l|XXXXXXXXXXXX}
& {\textsc{cmn}} & {\textsc{cym}} & {\textsc{eng}} & {\textsc{est}} & {\textsc{fin}} & {\textsc{fra}} & {\textsc{heb}} & {\textsc{pol}} & {\textsc{rus}} & {\textsc{spa}} & {\textsc{swa}} & {\textsc{yue}}  \\
\hline
{\textsc{cmn}} & \cellcolor{Gray}{} & {.076} & {.348} & {.139} & {.154} & {.392} & {.190} & {.207} & {.227} & {.300} & {.049} & {.484} \\
{\textsc{cym}} &  {.041} & \cellcolor{Gray}{} & {.087} & {.017} & {.049} & {.095} & {.033} & {.072} & {.085} & {.089} & {.002} & {.083} \\
{\textsc{eng}} &  {.565} & {.004} & \cellcolor{Gray}{} & {.168} & {.159} & {.401} & {.171} & {.182} & {.236} & {.309} & {.014} & {.357} \\
{\textsc{est}} &  {.014} & {.097} & {.335} & \cellcolor{Gray}{} & {.143} & {.161} & {.100} & {.113} & {.083} & {.134} & {.025} & {.124} \\
{\textsc{fin}} &  {.049} & {.020} & {.542} & {.530} & \cellcolor{Gray}{} & {.195} & {.077} & {.110} & {.111} & {.157} & {.029} & {.167} \\
{\textsc{fra}} &  {.224} & {.015} & {.662} & {.559} & {.533} & \cellcolor{Gray}{} & {.191} & {.229} & {.297} & {.382} & {.038} & {.382} \\
{\textsc{heb}} &  {.202} & {.110} & {.516} & {.465} & {.445} & {.469} & \cellcolor{Gray}{} & {.095} & {.154} & {.181} & {.038} & {.185} \\
{\textsc{pol}} &  {.121} & {.028} & {.464} & {.415} & {.465} & {.534} & {.412} & \cellcolor{Gray}{} & {.139} & {.183} & {.013} & {.205} \\
{\textsc{rus}} &  {.032} & {.037} & {.511} & {.408} & {.476} & {.529} & {.430} & {.390} & \cellcolor{Gray}{} & {.248} & {.037} & {.226} \\
{\textsc{spa}} &  {.546} & {.048} & {.498} & {.450} & {.490} & {.600} & {.462} & {.398} & {.419} & \cellcolor{Gray}{} & {.055} & {.313} \\
{\textsc{swa}} &  {-.01} & {.116} & {.029} & {.006} & {.013} & {-.05} & {.033} & {.052} & {.035} & {.045} & \cellcolor{Gray}{} & {.043} \\
{\textsc{yue}} &  {.004} & {.047} & {.059} & {.004} & {.002} & {.059} & {.001} & {.074} & {.032} & {.089} & {-.02} & \cellcolor{Gray}{} \\
\bottomrule
\end{tabularx}
}
\vspace{-0.0mm}
\caption{Spearman's $\rho$ correlation scores on all 66 cross-lingual datasets. 1) The scores \textbf{below the main diagonal} are computed based on cross-lingual word embeddings (CLWEs) induced by aligning CC+Wiki \textsc{ft} in all languages (except for \textsc{yue} where we use Wiki \textsc{ft}) in a fully unsupervised way (i.e., without any bilingual supervision). We rely on a standard CLWE mapping-based (i.e., alignment) approach: \textsc{vecmap} \cite{Artetxe:2018acl}. 2) The scores \textbf{above the main diagonal} are computed by obtaining 768-dimensional word-level vectors from pretrained multilingual BERT (\textsc{m-bert}) following the procedure described in \S\ref{ss:mling_models}. For both fully unsupervised \textsc{vecmap} and \textsc{m-bert}, we report the results with unsupervised postprocessing enabled: all $2 \times 66$ reported scores are obtained using the \textsc{+abbt} (-10) variant.}
\vspace{-0.0mm}
\label{tab:xling-res}
\end{table}

\begin{figure}[t]
    \centering
    \begin{subfigure}[t]{0.5\textwidth}
        \centering
        \includegraphics[width=0.99\linewidth, trim={0 2mm 0 0}]{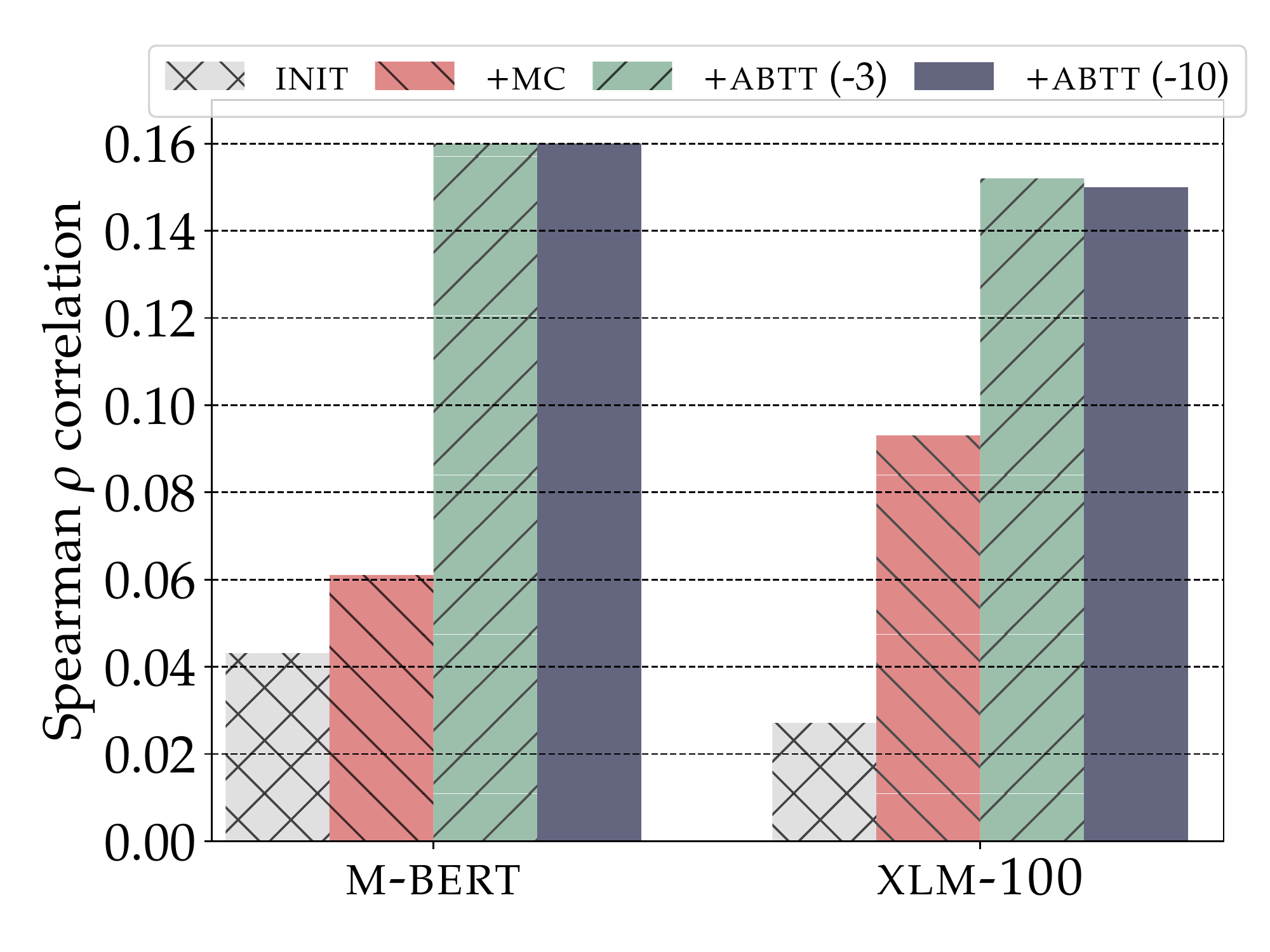}
        \caption{Average scores}
        \label{fig:avg-cross}
    \end{subfigure}%
    \vspace{0mm}
    \begin{subfigure}[t]{0.49\textwidth}
        \centering
        \includegraphics[width=0.98\linewidth,trim={0 2mm 0 0}]{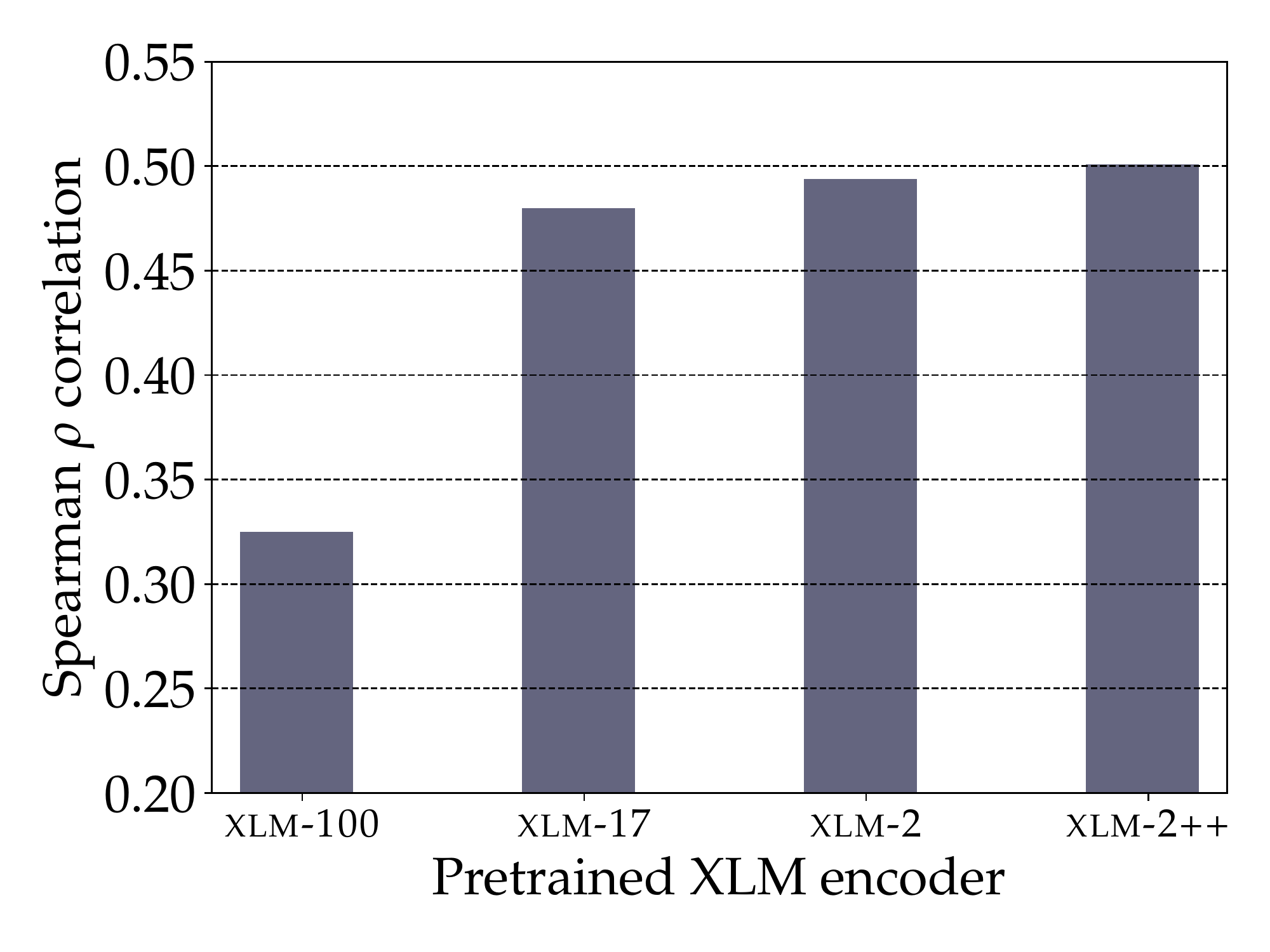}
        \caption{Scores on \textsc{eng-fra}}
        \label{fig:en-fr}
    \end{subfigure}
    \vspace{0mm}
    \caption{Further performance analyses of cross-lingual Multi-SimLex datasets. (a) Spearman's $\rho$ correlation scores averaged over all 66 cross-lingual Multi-SimLex datasets for two pretrained multilingual encoders (\textsc{m-bert} and \textsc{xlm}). The scores are obtained with different configurations that exclude (\textsc{init}) or enable unsupervised post-processing. (b) A comparison of various pretrained encoders available for the English-French language pair, see the main text for a short description of each benchmarked pretrained encoder.}
    \vspace{-0mm}
\label{fig:more-cross}
\end{figure}

\subsection{Results and Discussion}
\label{ss:xling-res}
\noindent \textit{Main Results and Differences across Language Pairs.} A summary of the results on the 66 cross-lingual Multi-SimLex datasets are provided in Table~\ref{tab:xling-res} and Figure~\ref{fig:avg-cross}. The findings confirm several interesting findings from our previous monolingual experiments (\S\ref{ss:mling-res}), and also corroborate several hypotheses and findings from prior work, now on a large sample of language pairs and for the task of cross-lingual semantic similarity. 

First, we observe that the fully unsupervised \textsc{vecmap} model, despite being the most robust fully unsupervised method at present, fails to produce a meaningful cross-lingual word vector space for a large number of language pairs (see the bottom triangle of Table~\ref{tab:xling-res}): many correlation scores are in fact no-correlation results, accentuating the problem of fully unsupervised cross-lingual learning for typologically diverse languages and with fewer amounts of monolingual data \cite{Vulic:2019we}. The scores are particularly low across the board for lower-resource languages such as Welsh and Kiswahili. It also seems that the lack of monolingual data is a larger problem than typological dissimilarity between language pairs, as we do observe reasonably high correlation scores with \textsc{vecmap} for language pairs such as \textsc{cmn-spa}, \textsc{heb-est}, and \textsc{rus-fin}. However, typological differences (e.g., morphological richness) still play an important role as we observe very low scores when pairing \textsc{cmn} with morphologically rich languages such \textsc{fin}, \textsc{est}, \textsc{pol}, and \textsc{rus}. Similar to prior work of \namecite{Vulic:2019we} and \namecite{doval2019onthe}, given the fact that unsupervised \textsc{vecmap} is the most robust unsupervised CLWE method at present \cite{Glavas:2019acl}, our results again question the usefulness of fully unsupervised approaches for a large number of languages, and call for further developments in the area of unsupervised and weakly supervised cross-lingual representation learning.

The scores of \textsc{m-bert} and \textsc{xlm-100}\footnote{The \textsc{xlm-100} scores are not reported for brevity; they largely follow the patterns observed with \textsc{m-bert}. The aggregated scores between the two encoders are also very similar as indicated by Figure~\ref{fig:avg-cross}.} lead to similar conclusions as in the monolingual settings. Reasonable correlation scores are achieved only for a small subset of resource-rich language pairs (e.g., \textsc{eng}, \textsc{fra}, \textsc{spa}, \textsc{cmn}) which dominate the multilingual \textsc{m-bert} training. Interestingly, the scores indicate a much higher performance of language pairs where \textsc{yue} is one of the languages when we use \textsc{m-bert} instead of \textsc{vecmap}. This boils down again to the fact that \textsc{yue}, due to its specific language script, has a good representation of its words and subwords in the shared \textsc{m-bert} vocabulary. At the same time, a reliable \textsc{vecmap} mapping between \textsc{yue} and other languages cannot be found due to a small monolingual \textsc{yue} corpus. In cases when \textsc{vecmap} does not yield a degenerate cross-lingual vector space starting from two monolingual ones, the final correlation scores seem substantially higher than the ones obtained by the single massively multilingual \textsc{m-bert} model.

Finally, the results in Figure~\ref{fig:avg-cross} again verify the usefulness of unsupervised post-processing also in cross-lingual settings. We observe improved performance with both \textsc{m-bert} and \textsc{xlm-100} when mean centering (\textsc{+mc}) is applied, and further gains can be achieved by using \textsc{abtt} on the mean-centered vector spaces. A similar finding also holds for static cross-lingual word embeddings\footnote{Note that \textsc{vecmap} does mean centering by default as one of its preprocessing steps prior to learning the mapping function \cite{Artetxe:2018acl,Vulic:2019we}.}, where applying \textsc{abbt} (-10) yields higher scores on 61/66 language pairs.

\vspace{1.6mm}
\noindent \textit{Fully Unsupervised vs. Weakly Supervised Cross-Lingual Embeddings.}
The results in Table~\ref{tab:xling-res} indicate that fully unsupervised cross-lingual learning fails for a large number of language pairs. However, recent work \cite{Vulic:2019we} has noted that these sub-optimal non-alignment solutions with the \textsc{unsuper} model can be avoided by relying on (weak) cross-lingual supervision spanning only several thousands or even hundreds of word translation pairs. Therefore, we examine 1) if we can further improve the results on cross-lingual Multi-SimLex resorting to (at least some) cross-lingual supervision for resource-rich language pairs; and 2) if such available word-level supervision can also be useful for a range of languages which displayed near-zero performance in Table~\ref{tab:xling-res}. In other words, we test if recent ``tricks of the trade'' used in the rich literature on CLWE learning reflect in gains on cross-lingual Multi-SimLex datasets.

First, we reassess the findings established on the bilingual lexicon induction task \cite{Sogaard:2018acl,Vulic:2019we}: using at least some cross-lingual supervision is always beneficial compared to using no supervision at all. We report improvements over the \textsc{unsuper} model for all 10 language pairs in Table~\ref{tab:xling-variants}, even though the \textsc{unsuper} method initially produced strong correlation scores. The importance of self-learning increases with decreasing available seed dictionary size, and the \textsc{+sl} model always outperforms \textsc{unsuper} with 1k seed pairs; we observe the same patterns also with even smaller dictionary sizes than reported in Table~\ref{tab:xling-variants} (250 and 500 seed pairs). Along the same line, the results in Table~\ref{tab:xling-variants2} indicate that at least some supervision is crucial for the success of static CLWEs on resource-leaner language pairs. We note substantial improvements on all language pairs; in fact, the \textsc{vecmap} model is able to learn a more reliable mapping starting from clean supervision. We again note large gains with self-learning.

\begin{table}[!t]
\def\arraystretch{0.99}
\centering
{\footnotesize
\begin{tabularx}{\linewidth}{l YYYYYYYYYY}
\toprule
\rowcolor{Gray}
{}  & {\textsc{cmn-eng}} & {\textsc{eng-fra}} & {\textsc{eng-spa}} & {\textsc{eng-rus}} & {\textsc{est-fin}} & {\textsc{est-heb}} & {\textsc{fin-heb}} & {\textsc{fra-spa}} & {\textsc{pol-rus}} & {\textsc{pol-spa}}\\
\cmidrule(lr){2-11}
{\textsc{unsuper}} & {.565} & {.662} & {.498} & {.511} & {.510} & {.465} & {.445} & {.600} & {.390} & {.398} \\ 
\hdashline
{\textsc{super} (1k)} & {.575} & {.602} & {.453} & {.376} & {.378} & {.363} & {.442} & {.588} & {.399} & {.406} \\
{\textsc{+sl} (1k)} & {.577} & {.703} & {.547} & {.548} & {\bf .591} & {.513} & {.488} & {.639} & {\bf .439} & {.456} \\
\hdashline
{\textsc{super} (5k)} & {\bf .587} & {.704} & {.542} & {.535} & {.518} & {.473} & {.585} & {.631} & {.455} & {.463} \\
{\textsc{+sl} (5k)} & {.581} & {\bf .707} & {\bf .548} & {\bf .551} & {.556} & {\bf .525} & {\bf .589} & {\bf .645} & {.432} & {\bf .476} \\
\bottomrule
\end{tabularx}
}%
\vspace{0mm}
\caption{Results on a selection of cross-lingual Multi-SimLex datasets where the fully unsupervised (\textsc{unsuper}) CLWE variant yields reasonable performance. We also show the results with supervised \textsc{vecmap} without self-learning (\textsc{super}) and with self-learning (\textsc{+sl}), with two seed dictionary sizes: 1k and 5k pairs; see \S\ref{ss:xling-models} for more detail. Highest scores for each language pair are in \textbf{bold}.}
\label{tab:xling-variants}
\end{table}

\begin{table}[!t]
\def\arraystretch{0.99}
\centering
{\footnotesize
\begin{tabularx}{\linewidth}{l YYYYYYY}
\toprule
\rowcolor{Gray}
{}  & {\textsc{cmn-fin}} & {\textsc{cmn-rus}} & {\textsc{cmn-yue}} & {\textsc{cym-fin}} & {\textsc{cym-fra}} & {\textsc{cym-pol}} & {\textsc{fin-swa}} \\
\cmidrule(lr){2-8}
{\textsc{unsuper}} & {.049} & {.032} & {.004} & {.020} & {.015} & {.028} & {.013} \\ 
\hdashline
{\textsc{super} (1k)} & {.410} & {.388} & {.372} & {.384} & {.475} & {.326} & {.206} \\ 
{\textsc{+sl} (1k)} & {\bf .590} & {\bf .537} & {\bf .458} & {\bf .471} & {\bf .578} & {\bf .380} & {\bf .264} \\ 
\bottomrule
\end{tabularx}
}%
\vspace{0mm}
\caption{Results on a selection of cross-lingual Multi-SimLex datasets where the fully unsupervised (\textsc{unsuper}) CLWE variant fails to learn a coherent shared cross-lingual space. See also the caption of Table~\ref{tab:xling-variants}.}
\label{tab:xling-variants2}
\end{table}

\vspace{1.6mm}
\noindent \textit{Multilingual vs. Bilingual Contextualized Embeddings.}
Similar to the monolingual settings, we also inspect if massively multilingual training in fact dilutes the knowledge necessary for cross-lingual reasoning on a particular language pair. Therefore, we compare the 100-language \textsc{xlm-100} model with i) a variant of the same model trained on a smaller set of 17 languages (\textsc{xlm-17}); ii) a variant of the same model trained specifically for the particular language pair (\textsc{xlm-2}); and iii) a variant of the bilingual \textsc{xlm-2} model that also leverages bilingual knowledge from parallel data during joint training (\textsc{xlm-2++}). We again use the pretrained models made available by \namecite{Conneau:2019nips}, and we refer to the original work for further technical details. 

The results are summarized in Figure~\ref{fig:en-fr}, and they confirm the intuition that massively multilingual pretraining can damage performance even on resource-rich languages and language pairs. We observe a steep rise in performance when the multilingual model is trained on a much smaller set of languages (17 versus 100), and further improvements can be achieved by training a dedicated bilingual model. Finally, leveraging bilingual parallel data seems to offer additional slight gains, but a tiny difference between \textsc{xlm-2} and \textsc{xlm-2++} also suggests that this rich bilingual information is not used in the optimal way within the \textsc{xlm} architecture for semantic similarity.

In summary, these results indicate that, in order to improve performance in cross-lingual transfer tasks, more work should be invested into 1) pretraining dedicated language pair-specific models, and 2) creative ways of leveraging available cross-lingual supervision (e.g., word translation pairs, parallel or comparable corpora) \cite{Liu:2019conll,Wu:2019arxiv,Cao:2020iclr} with pretraining paradigms such as \textsc{bert} and \textsc{xlm}. Using such cross-lingual supervision could lead to similar benefits as indicated by the results obtained with static cross-lingual word embeddings (see Table~\ref{tab:xling-variants} and Table~\ref{tab:xling-variants2}). We believe that Multi-SimLex can serve as a valuable means to track and guide future progress in this research area.

\section{Conclusion and Future Work}
\label{s:conclusion}

We have presented Multi-SimLex, a resource containing
human judgments on the semantic similarity of word pairs for 12 monolingual and 66 cross-lingual datasets. The languages covered are typologically diverse and include also under-resourced ones, such as Welsh and Kiswahili. The resource covers an unprecedented amount of 1,888 word pairs, carefully balanced according to their similarity score, frequency, concreteness, part-of-speech class, and lexical field. In addition to Multi-Simlex, we release the detailed protocol we followed to create this resource. We hope that our consistent guidelines will encourage researchers to translate and annotate Multi-Simlex -style datasets for additional languages. This can help and create a hugely valuable, large-scale semantic resource for multilingual NLP research.  

The core Multi-SimLex we release with this paper already enables researchers to carry out novel linguistic analysis as well as establishes a benchmark for evaluating representation learning models. Based on our preliminary analyses, we found that speakers of closely related languages tend to express equivalent similarity judgments. In particular, geographical proximity seems to play a greater role than family membership in determining the similarity of judgments across languages. Moreover, we tested several state-of-the-art word embedding models, both static and contextualized representations, as well as several (supervised and unsupervised) post-processing techniques, on the newly released Multi-SimLex. This enables future endeavors to improve multilingual representation learning with challenging baselines. In addition, our results provide several important insights for research on both monolingual and cross-lingual word representations: 

\vspace{1.4mm}
\noindent 1) Unsupervised post-processing techniques (mean centering, elimination of top principal components, adjusting similarity orders) are always beneficial independently of the language, although the combination leading to the best scores is language-specific and hence needs to be tuned. 

\vspace{1.4mm}
\noindent 2) Similarity rankings obtained from word embeddings for nouns are better aligned with human judgments than all the other part-of-speech classes considered here (verbs, adjectives, and, for the first time, adverbs). This confirms previous generalizations based on experiments on English.

\vspace{1.4mm}
\noindent 3) The factor having the greatest impact on the quality of word representations is the availability of raw texts to train them in the first place, rather than language properties (such as family, geographical area, typological features).

\vspace{1.4mm}
\noindent 4) Massively multilingual pretrained encoders such as \textsc{m-bert} \citep{devlin2018bert} and \textsc{xlm-100} \citep{Conneau:2019nips} fare quite poorly on our benchmark, whereas pretrained encoders dedicated to a single language are more competitive with static word embeddings such as fastText \citep{Bojanowski:2017tacl}. Moreover, for language-specific encoders, parameter reduction techniques reduce performance only marginally.

\vspace{1.4mm}
\noindent 5) Techniques to inject clean lexical semantic knowledge from external resources into distributional word representations were proven to be effective in emphasizing the relation of semantic similarity. In particular, methods capable of transferring such knowledge from resource-rich to resource-lean languages \citep{Ponti:2019emnlp} increased the correlation with human judgments for most languages, except for those with limited unlabelled data.

\vspace{1.4mm}
Future work can expand our preliminary, yet large-scale study on the ability of pretrained encoders to reason over word-level semantic similarity in different languages. For instance, we have highlighted how sharing the same encoder parameters across multiple languages may harm performance. However, it remains unclear if, and to what extent, the input language embeddings present in \textsc{xlm-100} but absent in \textsc{m-bert} help mitigate this issue. In addition, pretrained language embeddings can be obtained both from typological databases \citep{littell2017uriel} and from neural architectures \citep{malaviya2017learning}. Plugging these embeddings into the encoders in lieu of embeddings trained end-to-end as suggested by prior work \citep{Tsvetkov-2016,Ammar:2016tacl,ponti2019towards} might extend the coverage to more resource-lean languages.

Another important follow-up analysis might involve the comparison of the performance of representation learning models on multilingual datasets for both word-level semantic similarity and sentence-level Natural Language Understanding. In particular, Multi-SimLex fills a gap in available resources for multilingual NLP and might help understand how lexical and compositional semantics interact if put alongside existing resources such as XNLI \cite{Conneau:2018emnlp} for natural language inference or PAWS-X \cite{Yang:2019emnlp} for cross-lingual paraphrase identification. Finally, the Multi-SimLex annotation could turn out to be a unique source of evidence to study the effects of polysemy in human judgments on semantic similarity: for equivalent word pairs in multiple languages, are the similarity scores affected by how many senses the two words (or multi-word expressions) incorporate?



In light of the success of initiatives like Universal Dependencies for multilingual treebanks, we hope that making Multi-SimLex and its guidelines available will encourage other researchers to expand our current sample of languages. We particularly encourage creation and submission of comparable Multi-SimLex datasets for under-resourced and typologically diverse languages in future work. In particular, we have made a Multi-Simlex community website available to facilitate easy creation, gathering, dissemination, and use of annotated datasets: \url{https://multisimlex.com/}.


\begin{acknowledgments}
This work is supported by the ERC Consolidator Grant LEXICAL: Lexical Acquisition Across Languages (no 648909). Thierry Poibeau is partly supported by a PRAIRIE 3IA Institute fellowship ("Investissements d'avenir" program, reference ANR-19-P3IA-0001). 
\end{acknowledgments}

\starttwocolumn
\bibliography{refs}

\end{document}